%% file: main.tex
\pdfoutput=1

\documentclass[11pt]{article}
\usepackage[protrusion=true,expansion=true]{microtype}

\usepackage{acl}

\usepackage{times}
\usepackage{latexsym}

\usepackage[T1]{fontenc}

\usepackage[utf8]{inputenc}

\usepackage{microtype}

\usepackage{inconsolata}

\usepackage{graphicx}
\usepackage{multirow}
\usepackage{subcaption}
\usepackage{enumitem}
\usepackage{xcolor}
\usepackage{setspace}
\usepackage{amsmath}
\usepackage{booktabs}
\usepackage{array}
\usepackage{mathtools}
\usepackage{amssymb}
\usepackage[misc]{ifsym}

\title{MuSC: Improving Complex Instruction Following with Multi-granularity Self-Contrastive Training}

\author{
    Hui Huang\textsuperscript{1}$^{*}$, 
    Jiaheng Liu\textsuperscript{2}$^{*}$, 
    Yancheng He\textsuperscript{1}$^{*}$, 
    Shilong Li\textsuperscript{3}, 
    Bing Xu\textsuperscript{1}, 
    Conghui Zhu\textsuperscript{1},
\\
    \textbf{Muyun Yang\textsuperscript{1}$^{\textsuperscript{\Letter}}$,} 
    \textbf{Tiejun Zhao\textsuperscript{1}}
\\
    \textsuperscript{1}Faculty of Computing, Harbin Institute of Technology, \textsuperscript{2}Nanjing University \\
    \textsuperscript{3}School of Artificial Intelligence, Beijing University of Posts and Telecommunications \\
    \texttt{huanghui@stu.hit.edu.cn, yangmuyun@hit.edu.cn}
}

\begin{document}
\maketitle
\let\oldthefootnote\thefootnote

\let\thefootnote\relax\footnotetext{$*$ Equal contribution. \Letter\:Corresponding Author.}
\let\thefootnote\relax\footnotetext{\textsuperscript{1}Codes are openly available at \url{https://github.com/HuihuiChyan/MuSC}.}
\let\thefootnote\oldthefootnote
\input{content/0_Abstract}
\input{content/1_Introduction}

\input{content/2_RelatedWork}
\input{content/3_Approach}
\input{content/4_Experiments}
\input{content/5_Analysis}
\input{content/6_Conclusion}

\bibliography{custom}
\input{content/8_Appendix}
\label{sec:appendix}
\end{document}

%% file: content/0_Abstract.tex
\begin{abstract}
Complex instruction-following with elaborate constraints is imperative for Large Language Models (LLMs). While existing methods have constructed data for complex instruction alignment, they all rely on a more advanced model, especially GPT-4, limiting their application. In this paper, we propose a Multi-granularity Self-Contrastive Training (MuSC) framework, to improve the complex instruction alignment without relying on a stronger model. Our method is conducted on both coarse and fine granularity. On coarse-granularity, we construct constraint-aware preference data based on instruction decomposition and recombination. On fine-granularity, we perform token-aware preference optimization with dynamic token-level supervision. Our method is evaluated on open-sourced models, and experiment results show our method achieves significant improvement on both complex and general instruction-following benchmarks, surpassing previous self-alignment methods\textsuperscript{1}.

\end{abstract}

%% file: content/1_Introduction.tex
\section{Introduction}

Large Language Models (LLMs) have made remarkable advancements and are being wildly applied across various domains \cite{zhao2024surveylargelanguagemodels,wu2024comparative,he2024chinese,liu2025comprehensive,he2024can,zhang2025codecriticbench}. The instruction-following ability is fundamental and important, as it enables LLMs to generate appropriate responses to given instructions and solve corresponding tasks~\cite{openai2024gpt4technicalreport}. 
While recent LLMs perform comparatively well on simple instructions, their response quality to complex instructions with elaborate constraints often falls under expectation, with some of the constraints omitted \cite{he2024complexsimpleenhancingmulticonstraint, jiang-etal-2024-followbench}, which hinders their application in more real-world complex scenarios.

To enhance the complex instruction following,
the core challenge 
is the scarcity of high-quality complex instruction data~\cite{lou2024large}. Most existing instruction datasets are constructed based on existing NLP datasets or question-answering websites with simple constraints \cite{camel2023,alpaca,OpenOrca,longpre2023flan}. To cope with the scarcity of complex instruction data, previous work such as Evol-Instruct \cite{xu2024wizardlm}, Conifer \cite{sun2024conifer}, and Self-Correct \cite{palmeira-ferraz-etal-2024-self-correction} have been proposed to construct complex instructions and responses. However, \textit{these methods typically rely on a high-performance proprietary model (e.g.,  GPT-4) to distill the complex instruction-following ability, which is expensive and can not be scaled up in real-world applications}.

\begin{figure}[t]
    \centering
    \includegraphics[width=0.48\textwidth]{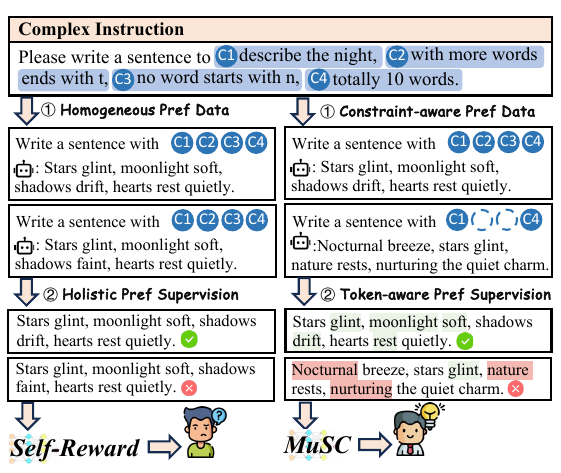}
    \caption{An illustrative comparison between our method and Self-Reward. Note that Self-Reward cannot create effective contrast for complex instruction-following, resulting in suboptimal optimization. }
    \label{fig: intro}
    \vspace{-1mm}
\end{figure}

Recently, the research community has paid attention to self-alignment, to break the data bottleneck without relying on a stronger model~\cite{wang2024comprehensivesurveyllmalignment, zhang2024surveyselfplaymethodsreinforcement}. Self-Reward \cite{yuan2024selfrewardinglanguagemodels} proposes to utilize the model itself to both generate responses and evaluate the results, which can be incorporated into DPO training. ISHEEP \cite{liang2024isheepselfalignmentllmscratch} proposes an automatic loop to self-assess and self-filter instruction data. \textit{Despite their effectiveness, these methods are targeted at general instruction-following ability. The self-alignment of complex instruction-following ability remains unexplored.}

In this paper, to address the above limitations, we propose a novel \textbf{Mu}lti-granularity \textbf{S}elf-\textbf{C}ontrastive Training framework (\textbf{MuSC}) in Figure~\ref{fig: intro}, which mainly comprises the following components:

1) \textbf{Coarse-grained Contrast: Constraint-aware Preference Data Construction}.  
To improve the model's comprehension of constraint-level distinctions, we construct preference pairs that reflect the disparities in constraint fulfillment. We achieve this by breaking down each complex instruction into atomic constraints and selectively omitting a subset to form negative instructions. The chosen response, derived from the original instruction, is paired with the rejected response, generated from the negative instruction, as a contrastive pair. Notably, no external models are utilized in this construction process.


2) \textbf{Fine-grained Contrast: Token-aware Preference Optimization}. For complex instructions, the responses often involve multiple tokens that contribute differently to fulfilling the instruction's constraints. Therefore, we introduce a token-aware optimization framework that integrates dynamic token-level weights based on the model’s confidence. By focusing on tokens that deviate from the constraints, this approach effectively identifies and corrects tokens where the model fails to satisfy the instruction’s requirements, leading to more contextually appropriate responses.

Moreover, we need to mention that our MuSC can be applied on both pre-existing complex instruction datasets, or newly generated instruction datasets created by data synthesis methods (e.g.,  Self-Instruct \cite{selfinstruct}). 

Our contribution can be summarized as follows:

\vspace{-2mm}

\begin{itemize}[itemsep=1mm, parsep=0pt]
    \item We propose a novel Multi-granularity Self-Contrastive Training (MuSC) framework, which creates effective contrast on both coarse and fine granularity, to enhance the complex instruction following abilities.
    \item For coarse-grained contrast, we construct constraint-aware preference data with instruction decomposition-recombination. For fine-grained contrast, we adopt dynamic token-level weight with confidence guidance for better preference optimization.
    \item We evaluate our framework on open-source LLMs, and achieve significant improvements on both complex and general instruction following benchmarks, without the help of a larger model or human supervision.
\end{itemize}

\vspace{-1mm}

%% file: content/2_RelatedWork.tex
\section{Related Work}
\noindent\textbf{Complex Instruction-Following}.
As one of the cores of LLM intelligence, how to improve the model's instruction-following capability is important. The earliest works, such as Alpaca \cite{alpaca}, Vicuna \cite{vicuna2023}, and Camel \cite{camel2023}, used instruction data generated by proprietary models to supervise fine-tuning of open-source models, 
significantly enhancing their instruction-following capabilities. However, these methods mainly focus on general instruction following, while complex instruction following still remains challenging.
To cope with this challenge, a lot of methods~\cite{yin2023dynosaur,lou2023muffin,he2024complexsimpleenhancingmulticonstraint,sun2024parrot,chen2024dog,dong2024self} have been proposed to construct complex instruction data. 
The earliest work is Evol-Instruct \cite{xu2024wizardlm}, which proposed to utilize GPT-4 to expand the instructions from both depth and width, thereby generating complex instructions and corresponding constraints. 
Conifer \cite{sun2024conifer} proposed a progressive learning strategy designed to help smaller models incrementally enhance their abilities.

\begin{figure*}[t]
    \centering
    \includegraphics[width=1\linewidth]{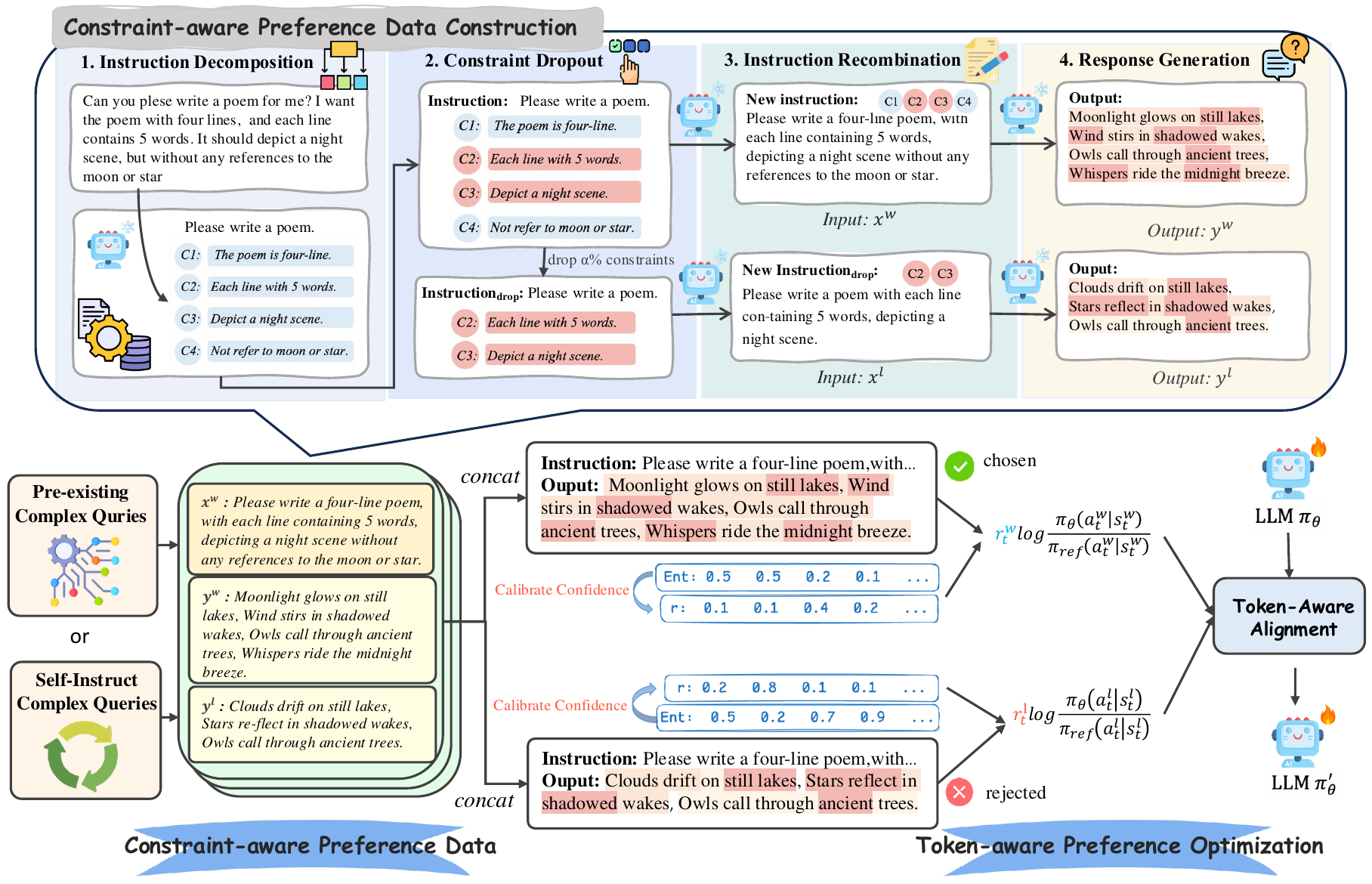}
    \caption{The pipeline of our proposed MuSC. The process starts with constraint-aware preference data construction, which includes instruction decomposition, constraint dropout, instruction recombination and response generation. Next, the token-aware DPO is performed based on calibrated confidence to achieve token-level alignment.}
    \label{fig: intro-2}
    \vspace{-1mm}
\end{figure*}

\noindent\textbf{Self-Alignment}.
Self-alignment refers to aligning the model to human preference without relying on a more advanced model or external supervision.
As an early study, Self-Rewarding \cite{yuan2024selfrewardinglanguagemodels} proposed the model itself to both generate responses and evaluate the results. 
Following this work, many works~\cite{liu-etal-2024-direct,chen2024dog,chen2024bootstrapping,pang2024self,meng2024simpo} are conducted to obtain supervision data by the model itself. Meta-Rewarding \cite{wu2024metarewardinglanguagemodelsselfimproving} advanced the concept by improving the model's instruction and evaluation capabilities simultaneously. 
\citet{liu-etal-2024-direct} employed diverse prompts to guide the model to generate various responses.
Despite the progress these methods have made, they all target general instruction following. For complex instructions with multiple constraints, the response will be lengthy and multi-facet, resulting in challenges for the self-evaluation process. 


%% file: content/3_Approach.tex
\section{Approach}
The pipeline of MuSC is shown in Figure~\ref{fig: intro-2}.
\subsection{Constraint-aware Preference Data}

Reinforcement-learning methods, such as PPO \cite{schulman2017proximalpolicyoptimizationalgorithms} and DPO \cite{rafailov2024directpreferenceoptimizationlanguage}, have achieved notable success in LLM optimization. Research has shown that learning from negative samples is significantly more efficient than learning solely from positive samples \cite{yang2024doesnegativesamplingmatter}. However, these methods are limited by the need for high-quality preference data, which is particularly scarce for complex instructions.


To construct effective preference data for complex instruction following, we propose a novel data construction method, with the following steps\footnote{Please refer to Appendix \ref{sec:impl} for implementation details.}:

\vspace{-2mm}

\begin{enumerate}[itemsep=1mm, parsep=0pt]
    \item Instruction Decomposition: A complex instruction is typically a combination of multiple atomic constraints. We decompose the complex instruction into individual atomic constraints, denoted as \texttt{Cons}.
    \item Constraint Dropout: From the decomposed constraints \texttt{Cons}, we randomly eliminate $\alpha\%$ of the constraints to form \texttt{Cons$_{drop}$}.
    \item Instruction Recombination: We recombine both the original and the dropped constraints \texttt{Cons} and \texttt{Cons$_{drop}$}, to create chosen and rejected instructions: \texttt{Ins} and \texttt{Ins$_{drop}$}.
    \item Response Generation: Based on \texttt{Ins} and \texttt{Ins$_{drop}$}, we generate the chosen response \texttt{Resp} and the rejected response \texttt{Resp$_{drop}$}.
\end{enumerate}

\vspace{-1mm}

Previous research has suggested that the construction of effective preference pairs for optimization is non-trivial \cite{ivison2024unpackingdpoppodisentangling}. Our data construction pipeline is guided by three principles:

\vspace{-1mm}

\begin{itemize}[itemsep=1mm, parsep=0pt]
    \item \textbf{Negativity}: The rejected response should deviate from the instruction by omitting some constraints. Our method generates the rejected instruction based on corrupted constraints, ensuring that the rejected response deviates from the original complex instruction.
    \item \textbf{Consistency}: The rejected response should reside within the model's decoding space \cite{guo2024directlanguagemodelalignment}. In our method, the rejected instruction is simply a recombination of the original instructions, ensuring the response falls within the decoding space, which is crucial for the optimization process.
    \item \textbf{Contrastiveness}: Chosen and rejected responses should be with a rational edit distance, to form an effective contrast \cite{jiang2024bridgingmodelingcorrelationspairwise}. By reconstructing both chosen and rejected instructions using the same method, we ensure that the derived samples do not deviate too far from each other.
\end{itemize}

\vspace{-2mm}

With constructed data satisfying both \textbf{negativity}, \textbf{consistency} and \textbf{contrastiveness}, we form a solid foundation for effective alignment. Moreover, our method does not require a stronger model or human supervision, ensuring its scalability.

Our self-construction method can be applied in different scenarios. On one hand, it can be directly applied on pre-existing complex instruction dataset. On the other hand, if there is no existing complex queries, we can adapt the Self-Instruct \cite{selfinstruct} method by first generating constraints and then generating instructions. In that case, the decomposition step can be omitted.

\subsection{Token-aware Preference Optimization}

A well-known issue with DPO is its uniform treatment of all tokens in both chosen and rejected examples \cite{wu2023finegrained, cao2024sparserewardsenhancingreinforcement, li20242ddposcalingdirectpreference}. However, different tokens within responses carry varying significance. Especially in scenarios involving complex instructions, the responses tend to be lengthy and multi-facet. On one hand, not all tokens in the rejected response are erroneous and should be disapproved. On the other hand, chosen response may also contain tokens that fail to meet specific constraints, therefore should not be unanimously approved.

Despite previous researchers have explored fine-grained supervision signals, the signals either come from a stronger model \cite{cao2024sparserewardsenhancingreinforcement, li20242ddposcalingdirectpreference} or human annotation \cite{wu2023finegrained, lightman2023let}. However, in our case, it is difficult for the model to provide accurate supervision for its own response, especially when dealing with multifaceted instructions and the evaluation is at token-level. Therefore, we propose Confidence-Guided Token-aware DPO, which obtains token-level supervision based on model confidence.

\subsubsection{Preliminary: Token-level DPO}
\label{preliminary}
Direct Preference Optimization (DPO) \cite{rafailov2024directpreferenceoptimizationlanguage} proposes a direct optimization objective that satisfies the optimal preference policy without using a reward model:



\vspace{-8mm}

\begin{align}
& \mathcal{L}_{DPO}(\pi_\theta; \pi_{ref}) = \nonumber\\
& -{E}_{\left(x, y^w, y^l \right) \sim \mathcal{D}} \left[ \log \sigma \left( \beta \log \frac{\pi_\theta(y_w \mid x)}{\pi_{ref}(y_w \mid x)} \right. \right. \nonumber\\
& \left. \left. - \beta \log \frac{\pi_\theta(y_l \mid x)}{\pi_{ref}(y_l \mid x)} \right) \right], 
\label{eq:dpo}
\end{align}

\vspace{-3mm}

\noindent where $\pi_\theta$ and $\pi_{ref}$ represent the policy model and the reference model, respectively. 

Subsequently, based on the theories of \citet{Levine2018ReinforcementLA}, \citet{rafailov2024rqlanguagemodel} derived the form of DPO in token-level Markov Decision Process\footnote{Please refer to Appendix \ref{app: prel} for more details.}, where dynamic weight can be easily integrated for different tokens\footnote{Please refer to Appendix \ref{app: change_beta} for the mathematical proof of the dynamic token weight in preference optimization}, with the loss function as follows:

\vspace{-8mm}

\begin{align}
& \mathcal{L}_{TDPO}(\pi_{\theta},D) = \nonumber \\
& -\mathbb{E}_{(\tau_w,\tau_l)\sim D} \log \sigma ( \beta \sum_{t=0}^{N-1} r_t^w \log \frac{\pi_{\theta}(\mathbf{a}_t^w|\mathbf{s}_t^w)}{\pi_{ref}(\mathbf{a}_t^w|\mathbf{s}_t^w)} \nonumber \\
& -\beta \sum_{t=0}^{M-1} r_t^l \log \frac{\pi_{\theta}(\mathbf{a}_t^l|\mathbf{s}_t^l)}{\pi_{ref}(\mathbf{a}_t^l|\mathbf{s}_t^l)}),
\label{eq: rdpo}
\end{align}

\vspace{-3mm}

\noindent where $\tau^w$ and $\tau^l$ represent the winning and losing trajectories, with $N$ and $M$ as the token numbers, and $r_t$ represents the weight for the $t$-th token.


\subsubsection{Calibrated Confidence as Token Weight}
\label{sec:entropy}

While Section ~\ref{preliminary} provide theoretical support for token-level DPO, it is non trivial to derive token-level supervision. In this work, we propose to use the calibrated confidence as supervision.

Given an instruction $x$, we obtain the entropy of probability distribution over target vocabulary of size $V$ at each decoding step as the weights:
\vspace{-2mm}
\begin{equation} 
\textrm{Ent}(y_t|x^{w}, \theta) = - \sum_{v=1}^{V}p(y_t^v)\mathrm{log}p(y_t^v),
\vspace{-1mm}
\end{equation}


\noindent where $p(y_t)$ represents the conditional distribution $p(y_t|x, y_{<t}, \theta)$, and $\theta$ represents model parameters. If the majority of the probability mass is concentrated on a limited number of vocabulary words, it indicates that the model is confident and the token is more likely to be aligned with the instruction~\cite{fomicheva2020unsupervised}. Conversely, if the probabilities resemble a uniform distribution, the resulting token is expected to be misaligned.


While there are other attributes that could also impact the confidence score (such as fluency), in this work, we want to focus our optimization on instruction alignment. Therefore, we apply calibration to the entropy to derive the token-level supervision for chosen and rejected samples respectively:

\vspace{-4mm}

\begin{align*} r_t =
\left\{  
  \begin{array}{ll}
    \textrm{Ent}(y_t|x^{w}, \theta) / \textrm{Ent}(y_t|x^{l}, \theta),\ \text{for}\ y_t\ \text{in}\ y^w, \\
    \textrm{Ent}(y_t|x^{l}, \theta) / \textrm{Ent}(y_t|x^{w}, \theta),\ \text{for}\ y_t\ \text{in}\ y^l,
  \end{array}
\right.
\end{align*}

\vspace{-4mm}
\begin{equation}
    r_t = \mathbf{min}(\Gamma, r_t),
\end{equation}
\vspace{-4mm}


\noindent where the chosen sample $(x^{w}, y^{w})$ refers to $(\texttt{Ins}, \texttt{Resp})$, and the rejected sample $(x^{l}, y^{l})$ refers to $(\texttt{Ins}_{drop}, \texttt{Resp}_{drop})$, and $\Gamma$ is an upper-bound to avoid extreme cases to disrupt training, and we set $\Gamma$ as 2 in this work.

The rationale is straightforward: for a given token in the response, if it exhibits high confidence under $x^{w}$ and low confidence under $x^{l}$, this suggests that the token aligns well with $x^{w}$ but does not fit $x^{l}$, potentially reflecting the dropped constraint. Therefore, the token requires a larger weight if it is in the rejected response (or a smaller weight if it is in the chosen response).


Calibrated confidence guided token-aware DPO allows for more targeted optimization, focusing on tokens that highlight the constraint mismatch on complex instructions, instead of unanimously optimizing all tokens, thereby improving the efficiency of complex instruction alignment.

%% file: content/4_Experiments.tex
\section{Experiments}
\label{sec:experiments}
\subsection{Set-up}
\paragraph{Models.}We  conduct experiments on two models: LLaMA-3-8B-Instruct \cite{dubey2024llama} and Qwen2-7B-Instruct \cite{yang2024qwen2}. Both models have undergone alignment to possess fundamental instruction-following ability.


\paragraph{Setting.} 
The experiments are carried out in two distinct settings:

\vspace{-2mm}

\begin{enumerate}[itemsep=1mm, parsep=0pt]
\item \textbf{Pre-Existing Instructions (PreInst):} We leverage pre-existing complex instructions as a starting point for the model. We randomly select 2,000 instructions from the dataset of WizardLM \cite{xu2024wizardlm}.
\item \textbf{Self-Generated Instructions (SelfInst):} In this setting, we generate instructions using the Self-Instruct method \cite{selfinstruct}, based on 10 high-quality samples from \citet{qin2024infobench} as in-context examples. Compared with \textbf{PreInst}, this setting is more challenging as we need to construct the complex instructions from scratch.
\end{enumerate}

\vspace{-2mm}


\begin{table*}[t]
\resizebox{1.0\textwidth}{!}{
\begin{tabular}{c|ccccccccc}
\toprule
\multirow{2}{*}{\textbf{Setting}} & \multicolumn{1}{c|}{\multirow{2}{*}{\textbf{Method}}} & \multicolumn{3}{c}{\textbf{CF-Bench}}         & \multicolumn{2}{c}{\textbf{FollowBench}} & \multicolumn{1}{c|}{\textbf{ComplexBench}} & \multicolumn{2}{c}{\textbf{AlpacaEval2}} \\
                                   & \multicolumn{1}{c|}{}                                 & \textbf{CSR}  & \textbf{ISR}  & \textbf{PSR}  & \textbf{HSR}        & \textbf{SSR}       & \multicolumn{1}{c|}{\textbf{Overall}}      & \textbf{LC (\%)}   & \textbf{Avg. Len}   \\
                                   \midrule
\multicolumn{10}{l}{\textit{Results on LLaMA-3-8B-Instruct}}                                                                                                                                                                                                             \\ 
\midrule
\multicolumn{2}{c|}{LLaMA-3-8B-Instruct}                                              & 0.64          & 0.24          & 0.34          & 62.39               & 73.07              & \multicolumn{1}{c|}{61.49}                 & 21.07              & 1702                \\
\midrule
\multirow{6}{*}{SelfInst}          & \multicolumn{1}{c|}{Self-Reward}                      & 0.65          & 0.26          & 0.35          & 61.20               & 72.22              & \multicolumn{1}{c|}{62.45}                 & 19.21              & 1824                \\
                                   & \multicolumn{1}{c|}{Self-Reward w/ BSM}               & 0.68          & 0.28          & 0.39          & 64.30               & 73.84              & \multicolumn{1}{c|}{64.13}                 & 19.03              & 1787                \\
                                   & \multicolumn{1}{c|}{Self-Reward w/ GPT-4}             & 0.66          & 0.25          & 0.37          & 62.18               & 73.34              & \multicolumn{1}{c|}{64.05}                 & 19.55              & 1767                \\ 
                                   & \multicolumn{1}{c|}{Self-Correct}                  & 0.52          & 0.20          & 0.27          & 54.38               & 67.19              & \multicolumn{1}{c|}{55.91}                 & 7.97               & 1919                \\ 
                                   & \multicolumn{1}{c|}{ISHEEP}                           & 0.60          & 0.29          & 0.40          & 62.77               & 72.86              & \multicolumn{1}{c|}{62.67}                 & 22.00              & 1707                \\ 
                                   & \multicolumn{1}{c|}{\textbf{MuSC}}                    & \textbf{0.70} & \textbf{0.32} & \textbf{0.44} & \textbf{66.71}      & \textbf{74.84}     & \multicolumn{1}{c|}{\textbf{65.98}}        & \textbf{23.87}     & 1708                \\
                                   \midrule
\multirow{7}{*}{PreInst}           & \multicolumn{1}{c|}{Self-Reward}                      & 0.66          & 0.27          & 0.37          & 60.88               & 72.17              & \multicolumn{1}{c|}{62.03}                 & 19.93              & 1789                \\
                                   & \multicolumn{1}{c|}{Self-Reward w/ BSM}               & 0.68          & 0.29          & 0.40          & 63.96               & 73.78              & \multicolumn{1}{c|}{64.3}                  & 20.98              & 1829                \\
                                   & \multicolumn{1}{c|}{Self-Reward w/ GPT-4}             & 0.66          & 0.26          & 0.37          & 64.02               & 73.26              & \multicolumn{1}{c|}{63.52}                 & 18.02              & 1804                \\ 
                                   & \multicolumn{1}{c|}{Self-Correct}                  & 0.60          & 0.23          & 0.32          & 60.11               & 70.94              & \multicolumn{1}{c|}{60.79}                 & 6.20               & 1593                \\ 
                                   & \multicolumn{1}{c|}{ISHEEP}                           & 0.67          & 0.29          & 0.40          & 63.54               & 73.21              & \multicolumn{1}{c|}{62.92}                 & 20.23              & 1703                \\ 
                                   & \multicolumn{1}{c|}{SFT}                              & 0.56          & 0.20          & 0.26          & 50.06               & 66.48              & \multicolumn{1}{c|}{53.93}                 & 10.00              & 1079                \\ 
                                   & \multicolumn{1}{c|}{\textbf{MuSC}}                    & \textbf{0.69} & \textbf{0.30} & \textbf{0.42} & \textbf{66.90}      & \textbf{75.11}     & \multicolumn{1}{c|}{\textbf{64.73}}        & \textbf{23.74}     & 1631                \\ 
                                   \midrule
\multicolumn{10}{l}{\textit{Results on Qwen2-7B-Instruct}}                                                                                                                                                                                                                    \\
\midrule
\multicolumn{2}{c|}{Qwen2-7B-Instruct}                                                     & 0.74          & 0.36          & 0.49          & 59.81               & 71.69              & \multicolumn{1}{c|}{67.24}                 & 15.53              & 1688                \\ \midrule
\multirow{5}{*}{SelfInst}          & \multicolumn{1}{c|}{Self-Reward}                      & 0.75          & 0.38          & 0.50          & 55.36               & 69.71              & \multicolumn{1}{c|}{66.98}                 & 16.81              & 1756                \\
                                   & \multicolumn{1}{c|}{Self-Reward w/ BSM}               & 0.75          & 0.38          & 0.50          & 57.83               & 70.53              & \multicolumn{1}{c|}{67.02}                 & 16.94              & 1710                \\ 
                                   & \multicolumn{1}{c|}{Self-Correct}                  & 0.67          & 0.28          & 0.38          & 51.98               & 67.89              & \multicolumn{1}{c|}{64.41}                 & 14.01              & 1497                \\ 
                                   & \multicolumn{1}{c|}{ISHEEP}                           & 0.76          & 0.40          & 0.52          & 57.01               & 69.88              & \multicolumn{1}{c|}{67.32}                 & 16.99              & 1619                \\ 
                                   & \multicolumn{1}{c|}{\textbf{MuSC}}                    & \textbf{0.78} & \textbf{0.42} & \textbf{0.54} & \textbf{62.60}      & \textbf{72.57}     & \multicolumn{1}{c|}{\textbf{69.39}}        & \textbf{20.08}     & 1595                \\ 
                                   \midrule
\multirow{6}{*}{PreInst}           & \multicolumn{1}{c|}{Self-Reward}                      & 0.75          & 0.37          & 0.49          & 56.45               & 70.00              & \multicolumn{1}{c|}{66.45}                 & 15.98              & 1796                \\
                                   & \multicolumn{1}{c|}{Self-Reward w/ BSM}               & 0.75          & 0.37          & 0.49          & 58.02               & 70.62              & \multicolumn{1}{c|}{67.43}                 & 17.17              & 1764                \\ 
                                   & \multicolumn{1}{c|}{Self-Correct}                  & 0.66          & 0.28          & 0.37          & 49.47               & 66.35              & \multicolumn{1}{c|}{64.32}                 & 14.46              & 1737                \\ 
                                   & \multicolumn{1}{c|}{ISHEEP}                           & 0.77          & 0.41          & 0.52          & 55.52               & 69.62              & \multicolumn{1}{c|}{67.13}                 & 16.52              & 1627                \\ 
                                   & \multicolumn{1}{c|}{SFT}                              & 0.72          & 0.35          & 0.46          & 47.36               & 64.67              & \multicolumn{1}{c|}{65.89}                 & 9.52               & 979                 \\
                                   & \multicolumn{1}{c|}{\textbf{MuSC}}                    & \textbf{0.79} & \textbf{0.44} & \textbf{0.55} & \textbf{62.73}      & \textbf{73.09}     & \multicolumn{1}{c|}{\textbf{70.00}}        & \textbf{20.29}     & 1613                \\
                                   \bottomrule
\end{tabular}}
\caption{Experiment results of different groups of methods on instruction following benchmarks. For more detailed results on each benchmark, please refer to Appendix \ref{sec:app-results}.}
\label{tab:main}
\end{table*}

\paragraph{Evaluation.} 
We mainly perform evaluations on three complex instruction-following benchmarks: \textbf{CFBench}~\cite{zhang2024cfbenchcomprehensiveconstraintsfollowingbenchmark}, \textbf{FollowBench}~\cite{jiang-etal-2024-followbench} and \textbf{ComplexBench} \cite{wen2024benchmarkingcomplexinstructionfollowingmultiple}. We also conduct evaluations on one general instruction benchmark: \textbf{AlpacaEval2}~\cite{dubois2024length}. Note that all benchmarks require GPT-4 for judgment, and we use GPT-4o-0513\footnote{\url{platform.openai.com/docs/models/gp##gpt-4o}} as the evaluator for all of them.


\paragraph{Baselines.} We mainly compared our method against the following self-alignment methods:

\vspace{-2mm}
\begin{itemize}[itemsep=1mm, parsep=0pt]
  \item \textbf{Self-Reward} \cite{yuan2024selfrewardinglanguagemodels}: This method leverages the model to first generate multiple responses and then construct rewards.
  \item \textbf{Self-Reward + BSM}: Based on Self-Reward, this method performs fine-grained evaluation based on BSM \cite{saha2024branchsolvemergeimproveslargelanguage}.
  \item \textbf{Self-Correct} \cite{palmeira-ferraz-etal-2024-self-correction}: This method generates initial output and then corrects it to construct preference data.
  \item \textbf{ISHEEP} \cite{liang2024isheepselfalignmentllmscratch}: This method self-creates additional instruction-output pair, which are filtered for supervised fine-tuning.
\end{itemize}
\vspace{-2mm}

\subsection{Main Results}

\begin{table*}[t]
\centering
\resizebox{1.0\textwidth}{!}{
\begin{tabular}{cc|cccccc|cc}
\toprule
\multirow{2}{*}{\textbf{Setting}} & \multirow{2}{*}{\textbf{Method}} & \multicolumn{3}{c}{\textbf{CF-Bench}}         & \multicolumn{2}{c}{\textbf{FollowBench}} & \textbf{ComplexBench} & \multicolumn{2}{c}{\textbf{AlpacaEval2}} \\
                                   &                                  & \textbf{CSR}  & \textbf{ISR}  & \textbf{PSR}  & \textbf{HSR}        & \textbf{SSR}       & \textbf{Overall}      & \textbf{LC (\%)}   & \textbf{Avg. Len}   \\ \midrule
\multicolumn{2}{c|}{LLaMA-3-8B-Ultrachat-200K}                         & 0.54          & 0.18          & 0.25          & 33.64               & 51.37              & 44.89                 & 5.94               & 861                 \\ \midrule
\multirow{4}{*}{PreInst}           & Self-Reward w/BSM                     & 0.58          & 0.2           & 0.28          & 39.24               & 56.82              & 53.69                 & 9.64               & 1099                \\
                                   & Self-Correct                     & 0.32          & 0.08          & 0.01          & 18.59               & 36.88              & 36.38                 & 3.47               & 575                 \\
                                   & ISHEEP                           & 0.57          & 0.19          & 0.26          & 42.22               & 58.87              & 52.95                 & 10.35              & 1283                \\ 
                                   & \textbf{MuSC}                    & \textbf{0.66} & \textbf{0.25} & \textbf{0.34} & \textbf{48.93}      & \textbf{61.54}     & \textbf{56.24}        & \textbf{11.13}     & 1112                \\ \midrule
\multicolumn{2}{c|}{LLaMA-3-8B-Tulu-330K}                              & 0.56          & 0.20          & 0.27          & 36.26               & 54.66              & 54.52                 & 6.00               & 992                 \\ \midrule
\multirow{4}{*}{PreInst}           & Self-Reward w/BSM                   & 0.60          & 0.22          & 0.30          & 41.71               & 58.68              & 55.84                 & 9.71               & 1307                \\
                                   & Self-Correct                     & 0.31          & 0.08          & 0.10          & 24.67               & 39.94              & 40.44                 & 3.21               & 623                 \\
                                   & ISHEEP                           & 0.61          & 0.22          & 0.30          & 43.16               & 59.21              & 58.20                 & 10.77              & 1402                \\ 
                                   & \textbf{MuSC}                    & \textbf{0.70} & \textbf{0.28} & \textbf{0.40} & \textbf{51.26}      & \textbf{63.94}     & \textbf{62.98}        & \textbf{13.20}     & 1341                \\ \bottomrule
\end{tabular}}
\caption{Experiment results of different methods on supervised fine-tuned models.}
\label{tab:sft}
\end{table*}

As demonstrated in Table \ref{tab:main}, our proposed MuSC achieves significant improvement across both complex and general instruction-following benchmarks. The improvement is consistent among different settings, verifying its scalability. By creating preference data with both constraint-aware and token-aware contrast, the model effectively learns to address all constraints lying in the instructions. 


The results of Self-Reward underperform our method, even with the help of branched evaluation \cite{saha2024branchsolvemergeimproveslargelanguage}. This is because of the limited evaluation capability of the model, especially when evaluating its own response to complex instructions. Moreover, as different responses generated from the same model to the same instruction typically do not vary significantly, it is difficult to create effective contrast samples with real negativity.

The improvement of I-SHEEP also underperforms, likely due to its reliance on supervised fine-tuning for optimization. Previous research also suggests that learning from negative samples is more effective than learning solely from positive ones \cite{yang2024doesnegativesamplingmatter}. The results of Self-Correct degrades a large margin, which might be due to the inability of the model for self-correction on complex instructions \cite{palmeira-ferraz-etal-2024-self-correction}.


On general instruction benchmarks, our method also achieves significant improvement. This aligns with the previous research, which suggests that the improvement on complex instruction-following is beneficial for the overall instruction-following ability \cite{xu2024wizardlm, elmadany-etal-2023-orca}.


%% file: content/5_Analysis.tex
\section{Analysis}

\begin{table}[t]
\centering
\resizebox{0.48\textwidth}{!}{
\begin{tabular}{c|ccc|cc}
\toprule
\multirow{2}{*}{\textbf{Method}} & \multicolumn{3}{c|}{\textbf{CF-Bench}}         & \multicolumn{2}{c}{\textbf{AlpacaEval2}}  \\
                                 & \textbf{CSR}  & \textbf{ISR}  & \textbf{PSR}  & \textbf{LC}     & \textbf{Len}    \\ \midrule
Baseline                         & 0.64          & 0.24          & 0.34          & 21.07                & 1702               \\ \midrule
Perplexity                    & 0.70          & 0.32          & 0.43          & 22.99                & 1744               \\
PMI                           & 0.69          & 0.29          & 0.41          & 21.92                & 1713               \\
KLDiv                         & 0.69          & 0.31          & 0.42          & 21.86                & 1686               \\ \midrule
\textbf{Entropy}              & \textbf{0.71} & \textbf{0.34} & \textbf{0.44} & \textbf{23.74}       & 1631               \\
w/o calib                     & 0.68          & 0.28          & 0.39          & 21.49                & 1735               \\\bottomrule
\end{tabular}}
\caption{Results of different confidence metrics as the fine-grained weight on LLaMA-3-8B-Instruct.}
\label{tab:confidence}
\end{table}

\begin{table}[t]
\resizebox{0.48\textwidth}{!}{
\begin{tabular}{cccccc}
\toprule
\multicolumn{1}{c|}{\multirow{2}{*}{\textbf{Method}}} & \multicolumn{3}{c|}{\textbf{CF-Bench}}                             & \multicolumn{2}{c}{\textbf{AlpacaEval2}} \\
\multicolumn{1}{c|}{}                                 & \textbf{CSR}  & \textbf{ISR}  & \multicolumn{1}{c|}{\textbf{PSR}}  & \textbf{LC}         & \textbf{Len}       \\ \midrule
\multicolumn{1}{c|}{Baseline}                         & 0.64          & 0.24          & \multicolumn{1}{c|}{0.34}          & 21.07               & 1702               \\ \midrule
\multicolumn{6}{l}{\textit{Results on SelfInst}}                                                                                                                      \\ \midrule
\multicolumn{1}{c|}{DPO$_{MuSC}$}                            & \textbf{0.70} & \textbf{0.32} & \multicolumn{1}{c|}{\textbf{0.44}} & \textbf{23.87}      & \textbf{1708}      \\ 
\multicolumn{1}{c|}{w/o fgct}                                & 0.68          & 0.28          & \multicolumn{1}{c|}{0.39}          & 21.49               & 1735        \\ \midrule
\multicolumn{1}{c|}{SimPO$_{MuSC}$}                          & \textbf{0.70} & \textbf{0.31} & \multicolumn{1}{c|}{\textbf{0.42}} & \textbf{22.92}      & \textbf{1716}      \\ 
\multicolumn{1}{c|}{w/o fgct}                                & 0.67          & 0.30          & \multicolumn{1}{c|}{0.40}          & 19.59               & 1637        \\ \midrule
\multicolumn{1}{c|}{IPO$_{MuSC}$}                            & \textbf{0.73} & \textbf{0.37} & \multicolumn{1}{c|}{\textbf{0.48}} & \textbf{23.18}      & \textbf{1650}      \\ 
\multicolumn{1}{c|}{w/o fgct}                                & 0.70          & 0.33          & \multicolumn{1}{c|}{0.44}          & 20.42               & 1686        \\ \midrule
\multicolumn{6}{l}{\textit{Results on PreInst}}                                                                                                                       \\ \midrule
\multicolumn{1}{c|}{DPO$_{MuSC}$}                            & \textbf{0.69} & \textbf{0.30} & \multicolumn{1}{c|}{\textbf{0.42}} & \textbf{23.74}      & \textbf{1631}      \\
\multicolumn{1}{c|}{w/o fgct}                                & 0.69          & 0.30          & \multicolumn{1}{c|}{0.41}          & 20.96               & 1671        \\ \midrule
\multicolumn{1}{c|}{SimPO$_{MuSC}$}                          & \textbf{0.69} & \textbf{0.31} & \multicolumn{1}{c|}{\textbf{0.42}} & \textbf{23.28}      & \textbf{1625}      \\
\multicolumn{1}{c|}{w/o fgct}                                & 0.67          & 0.31          & \multicolumn{1}{c|}{0.41}          & 22.02               & 1570        \\ \midrule
\multicolumn{1}{c|}{IPO$_{MuSC}$}                            & \textbf{0.68} & \textbf{0.32} & \multicolumn{1}{c|}{\textbf{0.44}} & \textbf{24.56}      & \textbf{1601}      \\
\multicolumn{1}{c|}{w/o fgct}                                & 0.68          & 0.30          & \multicolumn{1}{c|}{0.42}          & 22.04               & 1531        \\ \bottomrule
\end{tabular}}
\caption{Results of our method on different preference optimization methods on Llama-3-8B-Instruct.}
\label{tab:xpo}
\end{table}

\subsection{MuSC on SFT model}
In Section \ref{sec:experiments}, our main experiments are conducted on the Instruction-version models. To exclude the influence of an initial preference optimization process, we apply our method on SFT models.

Specifically, we selected two SFT-versions of LLaMA models, LLaMA-3-8B-UltraChat-200K and LLaMA-3-8B-Tulu-330K\footnote{\url{https://huggingface.co/Magpie-Align}}. Both models have gone through and only through SFT process on open-sourced datasets. As shown in Table \ref{tab:sft}, our proposed MuSC can improve both the complex and general instruction-following ability of SFT models by a large margin. Notice we only apply 2K samples when performing preference optimization, which is roughly 1\% of the amount of SFT data. This again verifies that learning from negative samples is comparatively more efficient than learning solely from positive samples.

\subsection{The Influence of Confidence Metrics}

Various confidence metrics have been established in the domain of LLM \cite{geng-etal-2024-confidence-survey}. This section aims to provide a comparison across different metrics as the token weight, under the framework of MuSC. We include the following metrics:

\vspace{-2mm}

\begin{itemize}[itemsep=1mm, parsep=0pt]
    \item \textbf{Perplexity}: The exponential of the negative log-likelihood of the token.
    \item \textbf{PMI}: Pointwise Mutual Information as defined in \citet{takayama-arase-2019-relevant}. 
    \item \textbf{KIDiv}: Kullback–Leibler divergence between the token probability distribution under chosen and rejected instructions.
\end{itemize}

\vspace{-2mm}

As shown in Table \ref{tab:confidence}, entropy-based token weight achieves the best result among all metrics, verifying its effectiveness. Both the perplexity and PMI-based score underperforms, as they only consider the probabilities of the selected tokens instead of the whole distribution, leading to biased evaluation\footnote{As the model would always select (one of) the tokens with the highest probability at each step, relying only on the the selected tokens for evaluation will result in overconfidence.}. KLDiv-based score also underperforms, this is because KLDiv is essentially a distance measurement instead of a confidence measurement, which is not adapted to our scenario.

We also experiment with removing the calibration proposed in Section \ref{sec:entropy}. As can be seen, calibration is important for the effectiveness of confidence-based fine-grained weight, as it can exclude other factors such as fluency, thereby focusing the contrast on instruction alignment.

\subsection{Can MuSC Scale to Other xPO Method?}
While our experiments primarily focus on the DPO method, the overall framework is not limited to a single preference optimization technique. Therefore, we extended our framework to two additional xPO methods: SimPO \cite{meng2024simpo} and IPO \cite{azar2024general}. We utilized the same constructed preference data and applied the entropy-based score as the token-level supervision\footnote{Please refer to Appendix \ref{sec:impl-dpo} for detailed implementation.}.
Note that for ``w/o fgct'', we just remove the fine-grained contrast in the MuSC method.
In Table \ref{tab:xpo}, our approach has shown consistent improvements across both SimPO and IPO, validating its scalability.

While IPO showed the best performance among different xPO methods, these performance differences were not statistically significant. Our method is not limited to DPO; instead, it's a general framework applicable to most xPO-style methods. Given the popularity of DPO, we primarily report our experiments based on DPO in this work.


\subsection{Is Fundamental Capability Harmed?}

\begin{table}[t]
\resizebox{0.48\textwidth}{!}{
\begin{tabular}{ccccc}
\toprule
\multicolumn{1}{c|}{}                                  &                                 &                                  & \multicolumn{1}{c|}{}                                     &                                \\
\multicolumn{1}{c|}{\multirow{-2}{*}{\textbf{Method}}} & \multirow{-2}{*}{\textbf{MMLU}} & \multirow{-2}{*}{\textbf{GSM8K}} & \multicolumn{1}{c|}{\multirow{-2}{*}{\textbf{HumanEval}}} & \multirow{-2}{*}{\textbf{Avg.}} \\ \midrule
\multicolumn{5}{l}{\textit{Results on Meta-LLaMA-3-8B-Instruct}}                                                                                                                                                         \\ \midrule
\multicolumn{1}{c|}{Baseline}                          & 68.29                           & 79.08                            & \multicolumn{1}{c|}{59.15}                                & 51.63                          \\ \midrule
\multicolumn{1}{c|}{MuSC$_{SelfInst}$}                            & 68.24                           & \textbf{79.23}                   & \multicolumn{1}{c|}{\textbf{62.20}}                       & \textbf{52.42}                 \\
\multicolumn{1}{c|}{w/o fgct}                      & {\color[HTML]{9B9B9B} 67.97}    & {\color[HTML]{9B9B9B} 78.62}     & \multicolumn{1}{c|}{{\color[HTML]{9B9B9B} 56.71}}         & {\color[HTML]{9B9B9B} 50.83}   \\ \midrule
\multicolumn{1}{c|}{MuSC$_{PreInst}$}                            & 68.17                           & 77.41                            & \multicolumn{1}{c|}{\textbf{62.80}}                       & \textbf{52.10}                 \\
\multicolumn{1}{c|}{w/o fgct}                      & 67.80                           & 77.71                            & \multicolumn{1}{c|}{\textbf{60.98}}                       & 51.62                          \\ \midrule
\multicolumn{5}{l}{\textit{Results on Qwen2-7B-Instruct}}                                                                                                                                                                \\ \midrule
\multicolumn{1}{c|}{Baseline}                          & 70.76                           & 83.09                            & \multicolumn{1}{c|}{75.61}                                & 57.37                          \\ \midrule
\multicolumn{1}{c|}{MuSC$_{SelfInst}$}                            & 70.63                           & \textbf{84.09}                   & \multicolumn{1}{c|}{75.61}                                & \textbf{57.58}                 \\
\multicolumn{1}{c|}{w/o fgct}                      & 70.81                           & 82.94                            & \multicolumn{1}{c|}{{\color[HTML]{9B9B9B} 71.34}}         & {\color[HTML]{9B9B9B} 56.27}   \\\midrule
\multicolumn{1}{c|}{MuSC$_{PreInst}$}                            & 70.46                           & \textbf{84.38}                   & \multicolumn{1}{c|}{75.78}                                & \textbf{57.66}                 \\ 
\multicolumn{1}{c|}{w/o fgct}                      & 70.63                           & \textbf{84.53}                   & \multicolumn{1}{c|}{{\color[HTML]{9B9B9B} 73.78}}         & {\color[HTML]{9B9B9B} 57.24}   \\\bottomrule
\end{tabular}}
\caption{Experiment results of our methods on fundamental capability benchmarks. Bolded results denote improvements, while grayed results denote degradation.}
\label{tab:fundamental}
\end{table}


Previous research have proposed that during the alignment process, the fundamental ability of model may suffer degradation due to alignment tax \cite{ouyang2022training}. Therefore, we evaluated our proposed method on three fundamental capability benchmarks: MMLU \cite{hendrycks2021measuring}, GSM8K \cite{cobbe2021trainingverifierssolvemath} and HumanEval \cite{chen2021evaluatinglargelanguagemodels}.


As shown in Table \ref{tab:fundamental}, while the results of naive MuSC may suffer slight degradation, the introduction of fine-grained contrast mitigates the degradation, which verifies the significance of token-level supervision. Under the scenario of complex instruction, the response is lengthy and should not be uniformly approved or disapproved. With fine-grained supervision, we focus the optimization on complex instruction alignment, thereby avoiding the disruption of other capabilities.

\subsection{The Variation of Statistical Indicators}

\begin{figure}[t]
\centering
\begin{subfigure}{0.48\textwidth}
    \includegraphics[width=\textwidth]{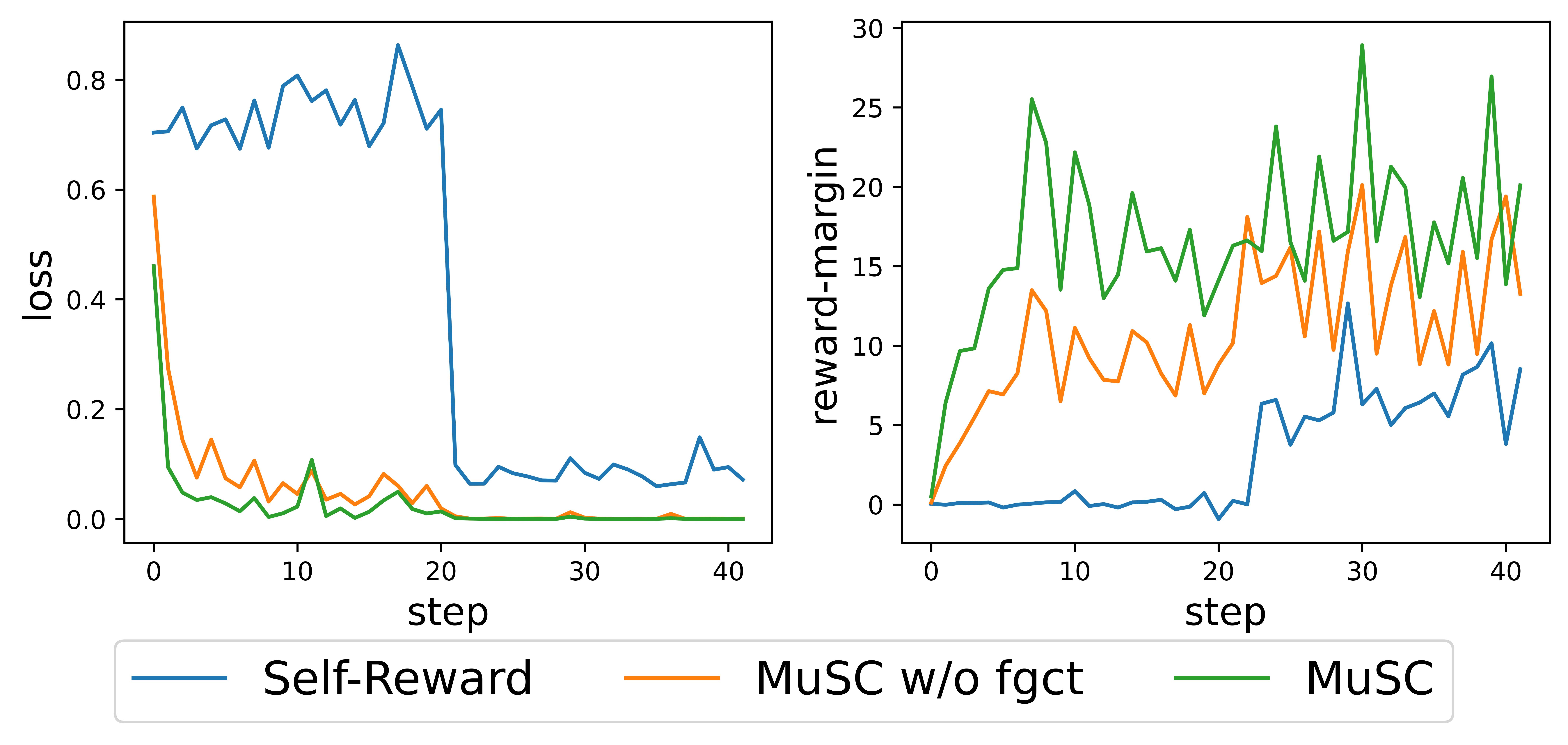}
    \caption{Experiment Results on Llama3-8B-Instruct.}
    \label{fig:first}
\end{subfigure}
\hfill
\begin{subfigure}{0.48\textwidth}
    \includegraphics[width=\textwidth]{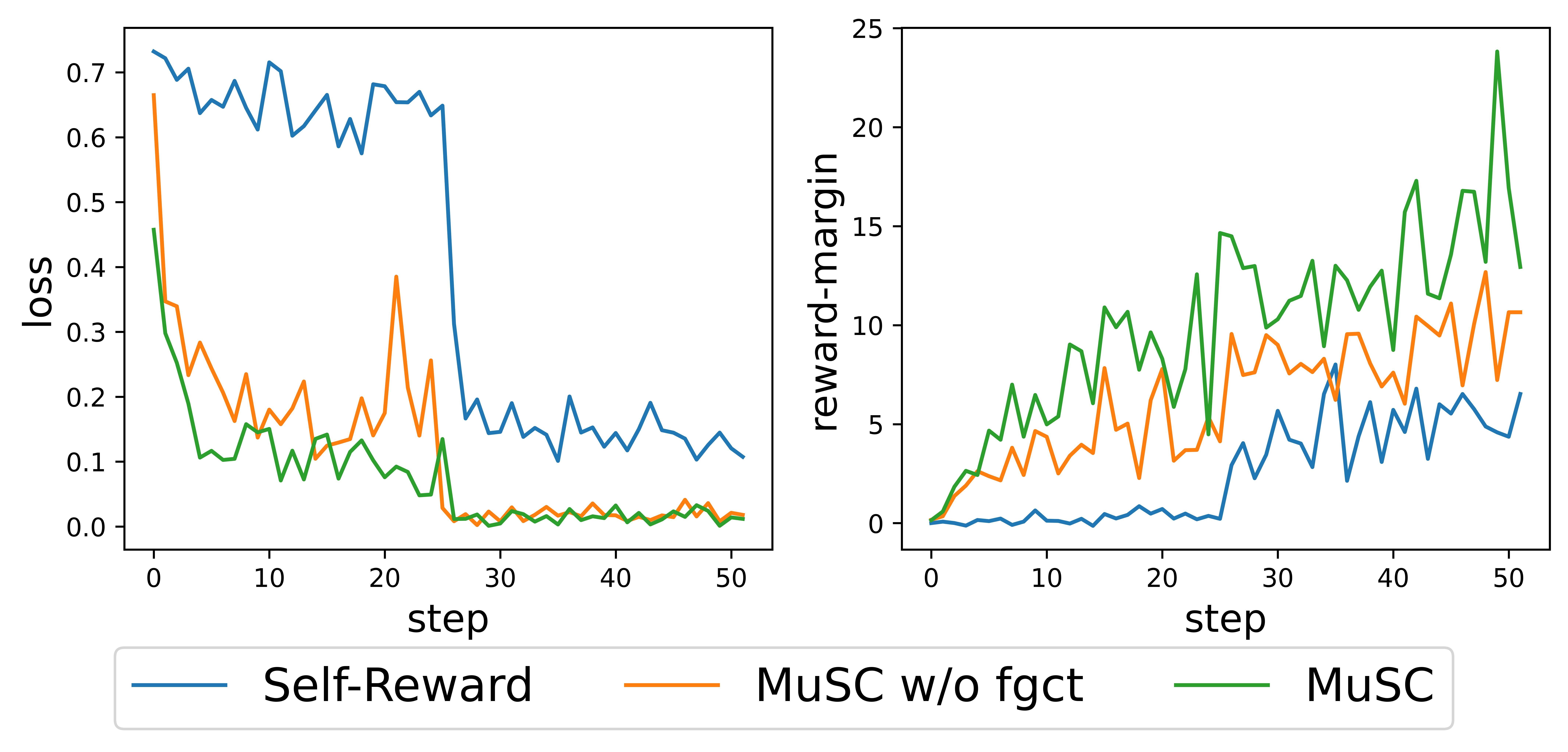}
    \caption{Experiment Results on Qwen2-7B-Instruct.}
    \label{fig:second}
\end{subfigure}        
\caption{Different statistical indicators during the training steps, upon different methods.}
\label{fig:indicator}
\end{figure}

As different methods start from the same group of instructions, training statistics can be comparable as a quality indicator. Therefore, we display the variation of both loss and reward margins between chosen and rejected samples on different methods. 


As shown in Figure \ref{fig:indicator}, Self-Reward presents both higher loss and lower reward margin during training. This is because there is too much noise between the chosen and rejected pairs, making the model unable to capture the contrast related to constraint alignment. Notice that both indicators start to change drastically at the 2nd epoch, which means the learned knowledge cannot transfer between different samples at the 1st epoch. On the other hand, the optimization based on MuSC converges faster and more smoothly, verifying the effectiveness of the contrast samples.

Comparing the indicators of MuSC with and without fine-grained supervision, it can also be noticed that with the introduction of fine-grained supervision, both indicators converge faster. The introduction of token-level supervision is a cheap yet effective method to improve xPO methods.

For the analysis of different instruction noising schemes and the visualization of token-level weight, please refer to Appendix \ref{app:noising} and \ref{sec:app-results}.

\subsection{Can MuSC Scale among Model Sizes}

\begin{table}[t]
\resizebox{0.48\textwidth}{!}{
\begin{tabular}{cccccc}
\hline
\multirow{2}{*}{\textbf{Model}} & \multicolumn{3}{c}{\textbf{CFBench}}          & \multicolumn{2}{c}{\textbf{AlpacaEval2}} \\
                                & \textbf{CSR}  & \textbf{ISR}  & \textbf{PSR}  & \textbf{LC}        & \textbf{Len}    \\ \hline
Qwen2.5-1.5B-Inst           & 0.59          & 0.21          & 0.28          & 10.10              & 1507                \\
+MuSC                           & \textbf{0.69} & \textbf{0.32} & \textbf{0.41} & \textbf{11.44}     & 1346                \\ \hline
Qwen2.5-3B-Inst             & 0.72          & 0.34          & 0.45          & 27.44              & 2188                \\
+MuSC                           & \textbf{0.76} & \textbf{0.41} & \textbf{0.52} & \textbf{30.64}     & 1873                \\ \hline
Qwen2.5-14B-Inst            & 0.82          & 0.49          & 0.60          & 46.05              & 1792                \\
+MuSC                           & \textbf{0.85} & \textbf{0.56} & \textbf{0.67} & \textbf{47.61}     & 1770                \\ \hline
Qwen2.5-32B-Inst            & 0.87          & 0.60          & 0.70          & 48.63              & 1785                \\
+MuSC                           & \textbf{0.89} & \textbf{0.65} & \textbf{0.75} & \textbf{49.34}     & 1772                \\ \hline
\end{tabular}}
\caption{Experiment results of MuSC on Qwen2.5-Instruct of different model sizes.}
\label{tab:scailibility}
\end{table}

To verify the scalability of our proposed method on models with larger sizes, we conducted extensive experiments on Qwen-2.5-Instruct models ranging from 1.5B to 32B parameters. As shown in Table \ref{tab:scailibility}, the results demonstrate consistent improvements across both smaller and larger models, confirming that our method is effective regardless of model size. This robust scalability is attributed to our approach being inherently self-improving and not requiring external models, which holds significant promise for advancing LLM capabilities.

%% file: content/6_Conclusion.tex
\section{Conclusion}

In this work, we propose a Multi-granularity Contrastive Training framework, to perform complex instruction alignment without the introduction of external supervision. Experiment results show our method achieves significant improvement on instruction alignment benchmarks, surpassing previous self-improvement methods by a large margin.
In the future, we will apply our MuSC on the improvement of other capabilities, such as long-form generation, multi-modal generation, etc.

\section*{Limitations}

Our work still has some limitations: 1) Due to time and resource limitation, we did not validate our method on larger models, such as LLaMA-70B. Validation on larger models could help to improve the credibility of our method. 2) We mainly relied on GPT-4 based LLM-as-a-Judge to evaluate the results. Despite it has been verified that GPT-4 based evaluation achieves high correlation with human evaluators \cite{zheng2023judging}, incorporating human evaluation would further improve the credibility of our methods. 3) We did not scale our method to PPO-based optimization methods, which are also wildly used in recent alignment practice. The application of our method on traditional RL methods could further improve its utility.

\section*{Ethical Considerations}
Since the aligned data is generated based on the Instruct version model, the chances of generating toxic content are low. To further validate this, we randomly selected 200 samples in total—100 samples for Llama-3-8B-Instruct and 100 samples for Qwen-2-7B-Instruct—and employed three graduate students to annotate them. The results showed no instances of toxic instructions or responses.

However, there is potential risk of knowledge forgetting related to safety alignment. To address this, we recommend incorporating additional safe alignment samples during MuSC training to reinforce the model's safety-related knowledge. We will include these recommendations in the ethical considerations section of the revised manuscript.

\section*{Acknowledgements}
This work is supported by National Natural Science Foundation of China (62276077, 62376075, 62376076), Jiangsu Science and Technology Major Project (BG2024031) and Nanjing University AI \& AI for Science Funding (2024300540).

%% file: content/8_Appendix.tex
\clearpage
\appendix
\onecolumn
\section{Implementation Details}
\subsection{Token-aware Preference Data Construction}
\label{sec:impl}
\begin{figure}[h]
    \centering
    \includegraphics[width=0.8\linewidth]{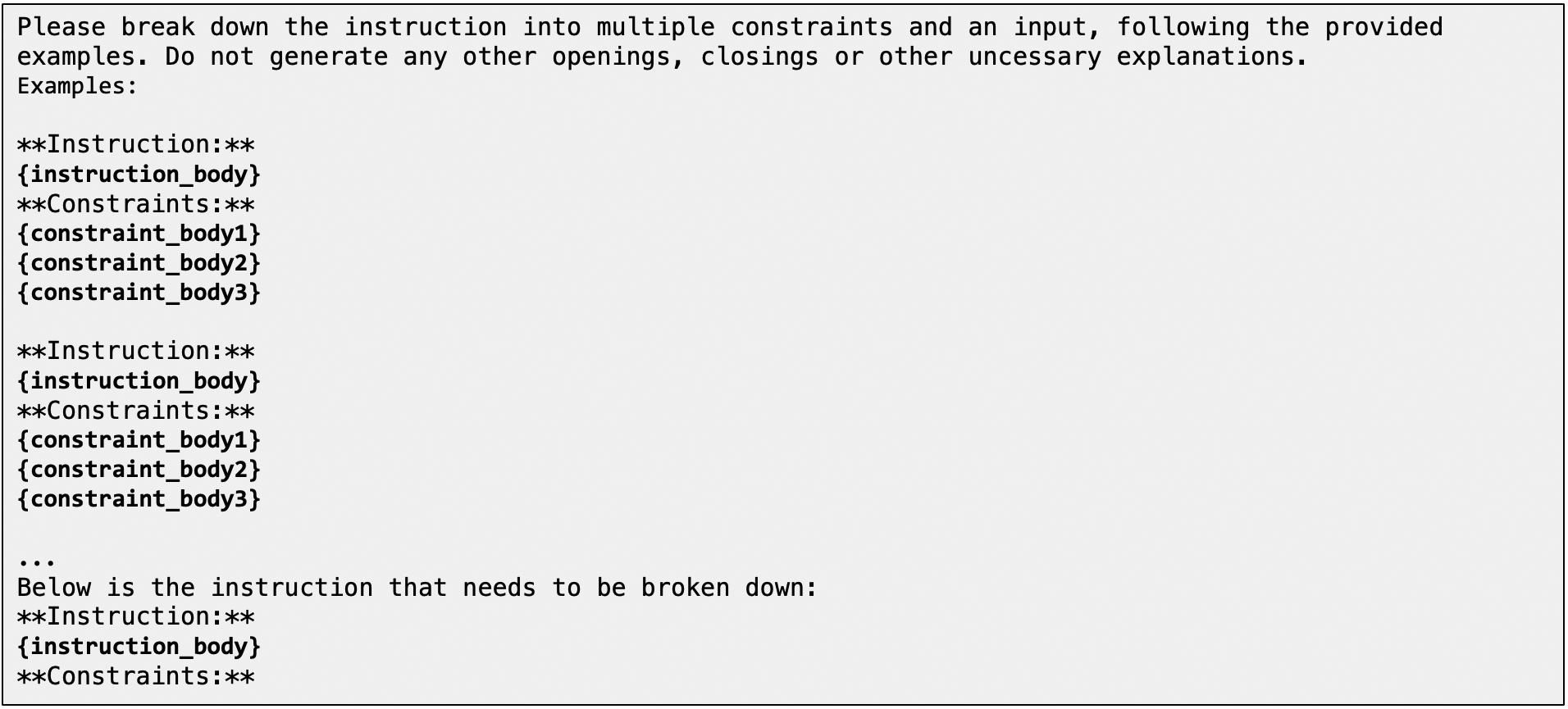}
    \caption{The prompt template used for instruction decomposition.}
    \label{fig: prompt-decom}
    \vspace{-1mm}
\end{figure}

\begin{figure}[h]
    \centering
    \includegraphics[width=0.8\linewidth]{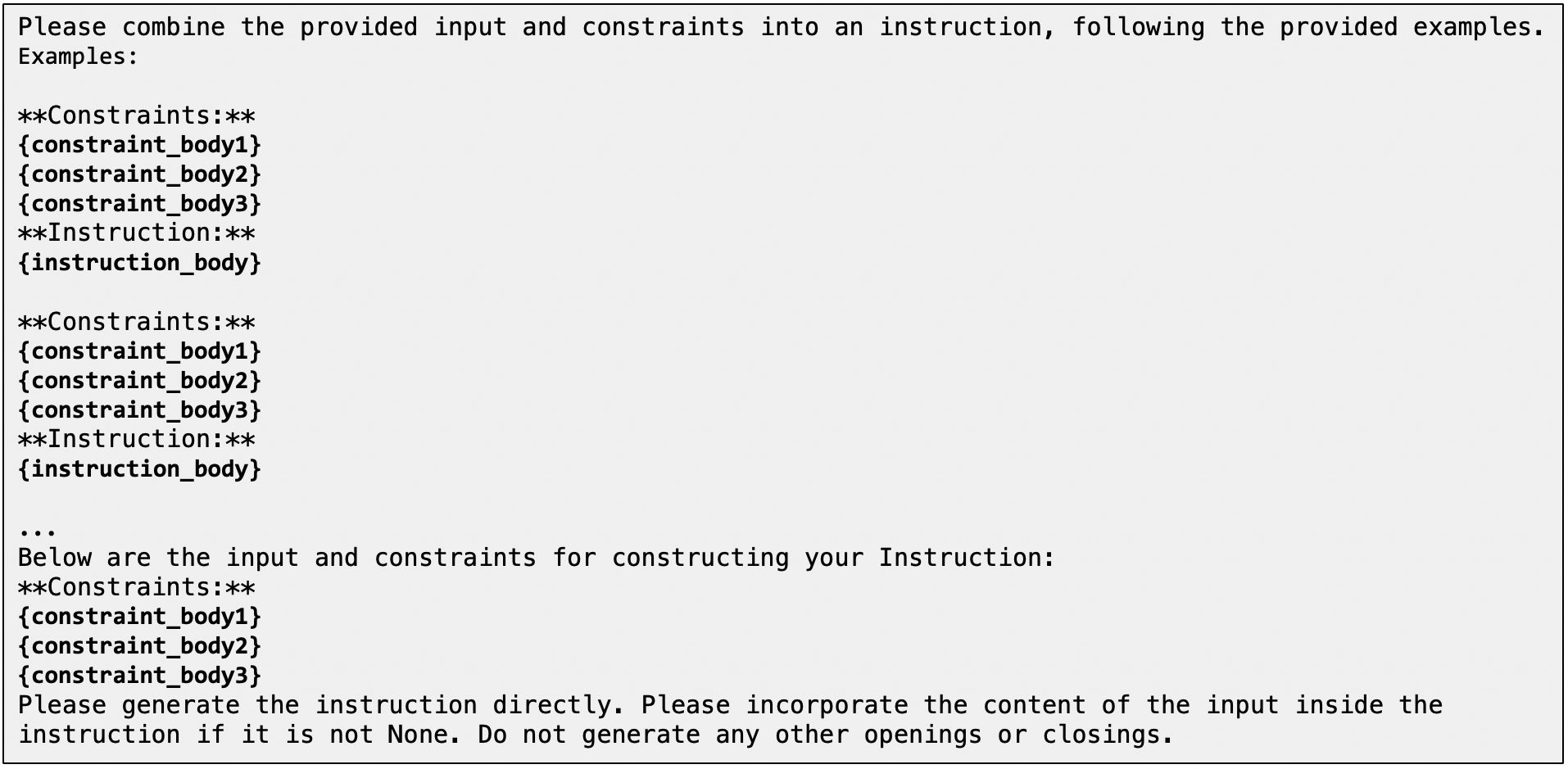}
    \caption{The prompt template used for constraint recombination.}
    \label{fig: prompt-recomb}
    \vspace{-1mm}
\end{figure}

\begin{figure}[h]
    \centering
    \includegraphics[width=0.8\linewidth]{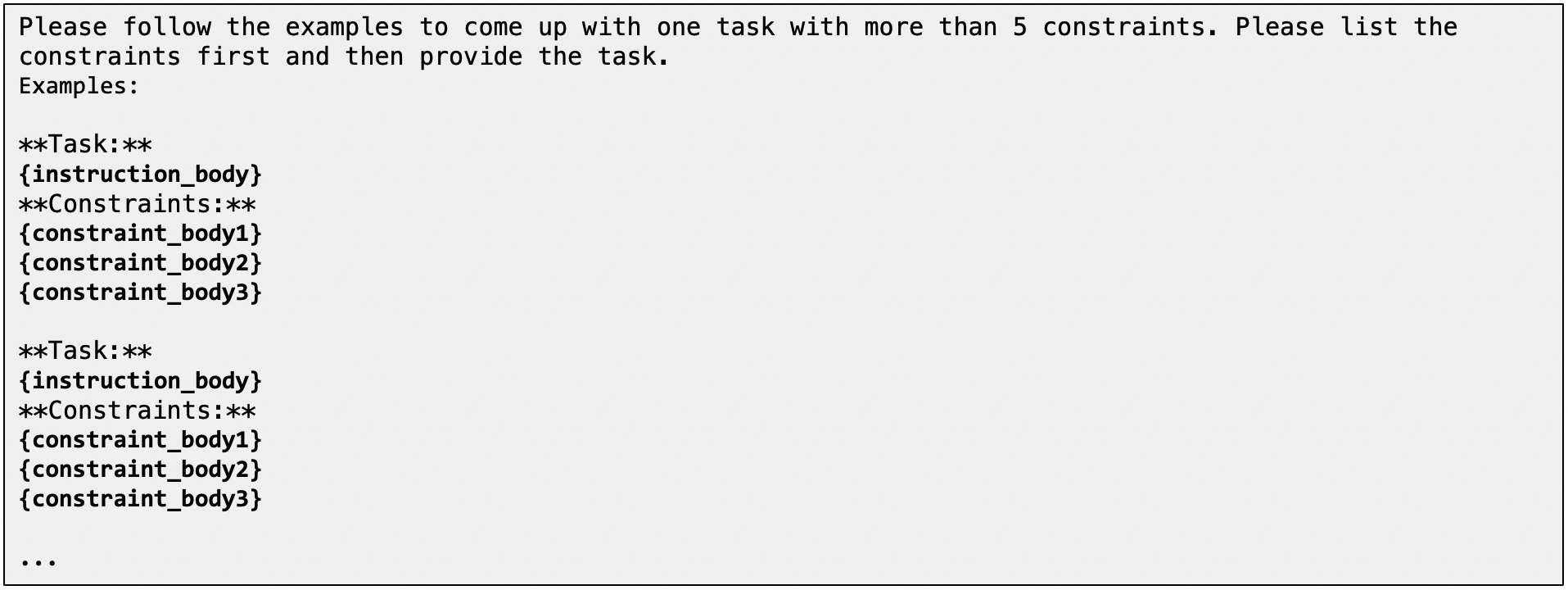}
    \caption{The prompt template used for self-instruct.}
    \label{fig: prompt-selfinst}
    \vspace{-1mm}
\end{figure}

\begin{figure}[h]
    \centering
    \includegraphics[width=0.8\linewidth]{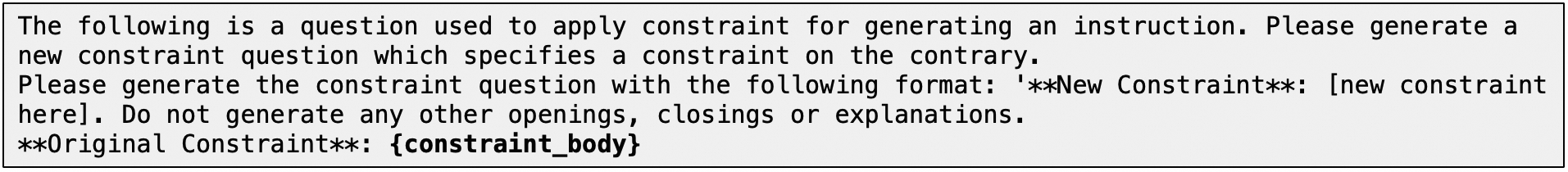}
    \caption{The prompt template used for constraint substitution.}
    \label{fig: prompt-sub}
    \vspace{-1mm}
\end{figure}

\begin{figure}[h]
    \centering
    \includegraphics[width=0.8\linewidth]{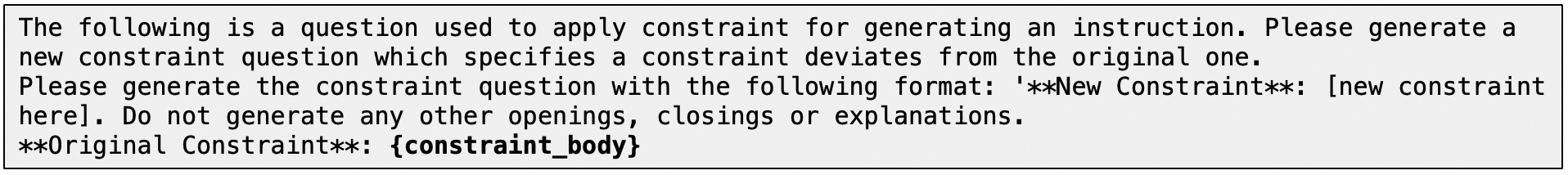}
    \caption{The prompt template used for constraint negation.}
    \label{fig: prompt-neg}
    \vspace{-1mm}
\end{figure}

For all models that used for preference data construction, we adopt the following prompts presented in Figure \ref{fig: prompt-decom}, \ref{fig: prompt-selfinst}, \ref{fig: prompt-recomb}, \ref{fig: prompt-sub}, \ref{fig: prompt-neg} and \ref{fig: prompt-sub}. We set the temperate as 0.5 for all steps to ensure diversity. To ensure the data quality, we filter instructions with less than three constraints and more than ten constraints. We also filter preference pairs with the same chosen and rejected responses. 

For constraint dropout, we set the dropout ratio $\alpha$ to 0.3 to ensure that negative examples are sufficiently negative, meanwhile not deviate too much from the positive sample. We avoid dropout on the first constraint, as it often establishes the foundation for the task, and dropping the first one would make the recombined instruction overly biased.

\subsection{Token-aware Preference Optimization}
\label{sec:impl-dpo}
Our experiments are based on Llama-Factory \cite{zheng2024llamafactory}, and we trained all models on 8 A100-80GB SXM GPUs. The \texttt{per\_device\_train\_batch\_size} was set to 1, \texttt{gradient\_accumulation\_steps} to 8, leading to an overal batch size as 64, and we used bfloat16 precision. The learning rate is set as 1e-06 with cosine decay,and each model is trained with 2 epochs. We set $\beta$ to 0.2 for all DPO-based experiments, $\beta$ as 3.0 and $\gamma$ as 1.0 for all SimPO-based experiments, $\beta$ as 1.0 for all IPO-based methods referring to the settings of \citet{meng2024simpo}. All of the final loss includes 0.1x of the SFT loss.

\section{The Influence of Noising Scheme}
\label{app:noising}

Previous work has proposed various noising strategies in contrastive training \cite{lai-etal-2021-saliency-based}. While we leverage Constraint-Dropout for negative sample generation, to make a fair comparison with other strategies, we implement the following strategies: 1) Constraint-Negate: Leverage the model to generate an opposite constraint. 2) Constraint-Substitute: Substitute the constraint with an unrelated constraint.

\begin{figure}[h]
\centering
\includegraphics[width=0.6\linewidth]{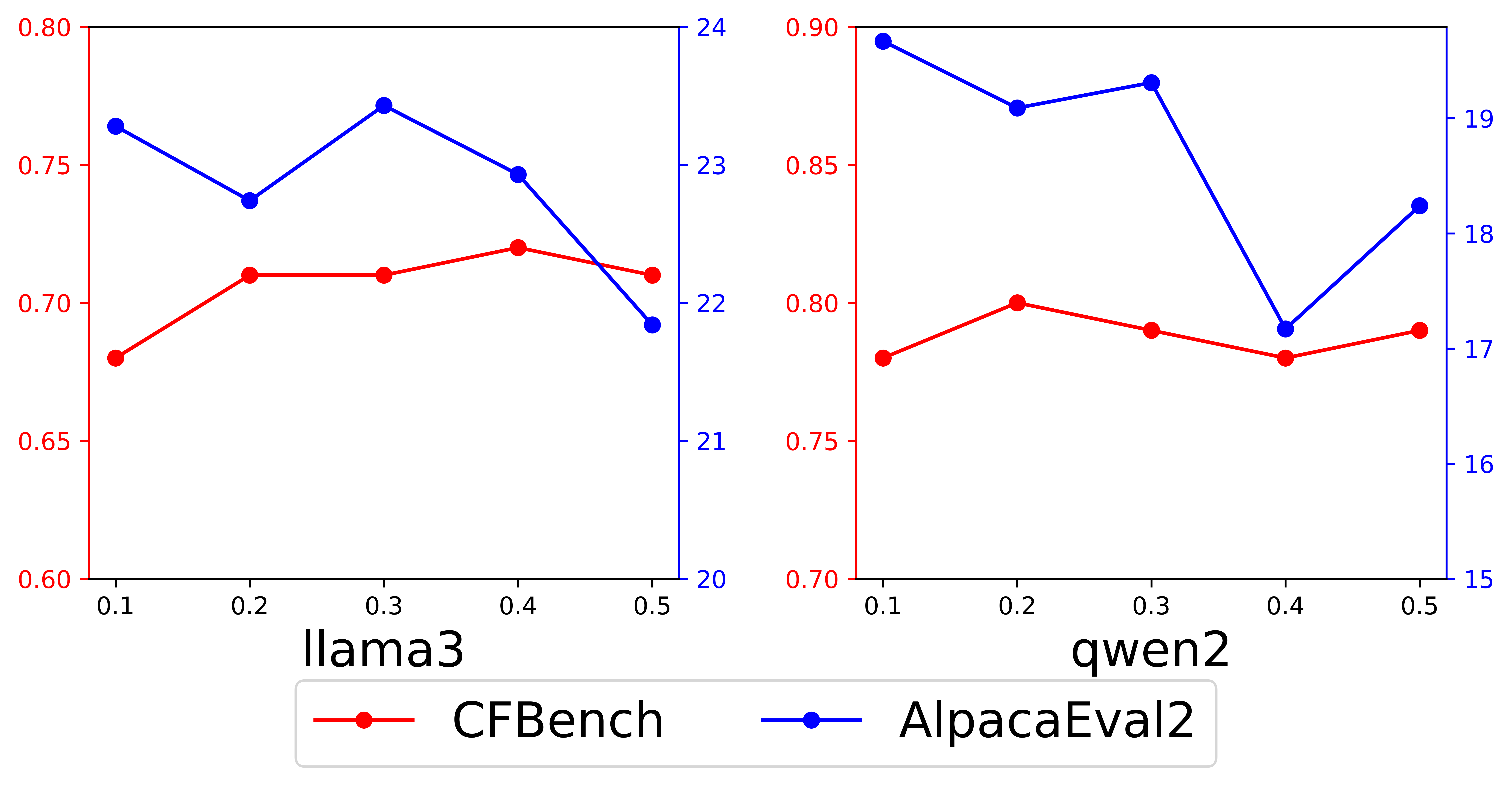}
\caption{The variation of results on CFBench and AlpacaEval2 with different dropout ratios.}
\label{fig:drop_ratio}
\end{figure}

As shown in Table \ref{tab:detail-noising}, both the negation and substitution applied on the constraints would lead to performance degradation. After a thoroughly inspect of the derived data, we realize that instructions derived from both dropout and negation would lead to instructions too far from the positive instruction, therefore the derived negative response would also deviate too much from the original instruction. An effective negative sample should fulfill both negativity, consistency and contrastiveness, and constrait-dropout is a simple yet effective method to achieve this goal.

We also provide the variation of the results on CF-Bench and AlpacaEval2 with different constraint dropout ratios. As shown in Figure \ref{fig:drop_ratio}, with the dropout ratio increased from 0.1 to 0.5, the results on CF-Bench firstly increases and then slightly decreases. On the other hand, the results on AlpacaEval2 declines a lot with a higher dropout ratio. This denotes that a suboptimal droout ratio is essential for the balance between complex instruction and general instruction following abilities, with lower ratio may decrease the effectiveness of general instruction alignment, while higher ratio may be harmful for complex instruction alignment. Finally, we set the constraint dropout ratio as 0.3 in all experiments.

\begin{table*}[tt]
\centering
\resizebox{1.0\textwidth}{!}{
\begin{tabular}{cc|ccccc|ccccc}
\toprule
\multirow{3}{*}{\textbf{Scenario}} & \multirow{3}{*}{\textbf{Method}} & \multicolumn{5}{c|}{\textbf{Meta-LLaMA-3-8B-Instruct}}                                    & \multicolumn{5}{c}{\textbf{Qwen-2-7B-Instruct}}                                          \\
                                   &                                  & \multicolumn{3}{c}{\textbf{CF-Bench}}         & \multicolumn{2}{c|}{\textbf{AlpacaEval2}} & \multicolumn{3}{c}{\textbf{CF-Bench}}         & \multicolumn{2}{c}{\textbf{AlpacaEval2}} \\
                                   &                                  & \textbf{CSR}  & \textbf{ISR}  & \textbf{PSR}  & \textbf{LC\%}      & \textbf{Avg.Len}     & \textbf{CSR}  & \textbf{ISR}  & \textbf{PSR}  & \textbf{LC\%}      & \textbf{Avg.Len}    \\ \midrule
\multirow{6}{*}{PreInst}           & baseline                         & 0.64          & 0.24          & 0.34          & 21.07              & 1702                 & 0.74          & 0.36          & 0.49          & 15.53              & 1688                \\ \cline{2-12} 
                                   & Constraint-Drop               & \textbf{0.71} & \textbf{0.34} & \textbf{0.45} & \textbf{23.43}     & 1682           & \textbf{0.79} & \textbf{0.43}  & \textbf{0.54}          & \textbf{19.31}     & 1675                \\
                                   & Constraint-Negate             & 0.68          & 0.28          & 0.39          & 18.94              & 1688                 & 0.75          & 0.37          & 0.50          & 17.82              & 1663                \\
                                   & Constraint-Substitute             & 0.68          & 0.28          & 0.40          & 20.48              & 1706                 & 0.76          & 0.39          & 0.51          & 19.05              & 1709                \\ \bottomrule
\end{tabular}}
\caption{Experiment results of different noising strategies on instruction following benchmarks.}
\label{tab:detail-noising}
\end{table*}

\section{Mathematical Derivations}
\subsection{Preliminary: DPO in the Token Level Marcov Decision Process}
\label{app: prel}


As demonstrated in \citet{rafailov2024rqlanguagemodel}, the Bradley-Terry preference model in token-level Marcov Decision Process (MDP) is:

\begin{equation}
p^*\left(\tau^w \succeq \tau^l\right)=\frac{\exp \left(\sum_{i=1}^N r\left(\mathbf{s}_i^w, \mathbf{a}_i^w\right)\right)}{\exp \left(\sum_{i=1}^N r\left(\mathbf{s}_i^w, \mathbf{a}_i^w\right)\right)+\exp \left(\sum_{i=1}^M r\left(\mathbf{s}_i^l, \mathbf{a}_i^l\right)\right)}
\label{eq: tdpo_bt}
\end{equation}

\label{app: tdpo}
The formula using the $Q$-function to measure the relationship between the current timestep and future returns:

\begin{equation}
Q^*(s_t, a_t) =
\begin{cases} 
r(s_t, a_t) + \beta \log \pi_{ref}(a_t | s_t) + V^*(s_{t+1}), & \text{if } s_{t+1} \text{ is not terminal} \\
r(s_t, a_t) + \beta \log \pi_{ref}(a_t | s_t), & \text{if } s_{t+1} \text{ is terminal}
\end{cases}
\label{eq: t_return}
\end{equation}

Derive the total reward obtained along the entire trajectory based on the above definitions:
\begin{align}
& \sum_{t=0}^{T-1} r(s_t, a_t)
 = \sum_{t=0}^{T-1} ( Q^*(s_t, a_t) - \beta \log \pi_{\text{ref}}(a_t | s_t) - V^*(s_{t+1}) )
\label{eq: r_sum}
\end{align}

Combining this with the fixed point solution of the optimal policy \cite{Ziebart2010ModelingPA, Levine2018ReinforcementLA}, we can further derive:
\begin{align}
\sum_{t=0}^{T-1} r(s_t, a_t)
& = Q^*(s_0, a_0) - \beta \log \pi_{ref}(a_0 | s_0) 
+ \sum_{t=1}^{T-1} ( Q^*(s_t, a_t) - V^*(s_t) - \beta \log \pi_{\text{ref}}(a_t | s_t) )
\\
& = Q^*(s_0, a_0) - \beta \log \pi_{ref}(a_0 | s_0) + \sum_{t=1}^{T-1} \beta \log \frac{\pi^*(a_t | s_t)}{\pi_{\text{ref}}(a_t | s_t)}
\\
& = V^*(s_0) + \sum_{t=0}^{T-1} \beta \log \frac{\pi^*(a_t | s_t)}{\pi_{\text{ref}}(a_t | s_t)}
\end{align}

By substituting the above result into Eq. \ref{eq: tdpo_bt}, we can eliminate $V^*(S_0)$ in the same way as removing the partition function in DPO, obtaining the Token-level BT model that conforms to the MDP:

\begin{equation}
p_{\pi^*}\left(\tau^w \succeq \tau^l\right)=\sigma\left(\sum_{t=0}^{N-1} \beta \log \frac{\pi^*\left(\mathbf{a}_t^w \mid \mathbf{s}_t^w\right)}{\pi_{\mathrm{ref}}\left(\mathbf{a}_t^w \mid \mathbf{s}_t^w\right)}-\sum_{t=0}^{M-1} \beta \log \frac{\pi^*\left(\mathbf{a}_t^l \mid \mathbf{s}_t^l\right)}{\pi_{\mathrm{ref}}\left(\mathbf{a}_t^l \mid \mathbf{s}_t^l\right)}\right)
\end{equation}

Thus, the Loss formulation of DPO at the Token level is:
\begin{equation}
\mathcal{L}\left(\pi_\theta, \mathcal{D}\right)=-\mathbb{E}_{\left(\tau_w, \tau_l\right) \sim \mathcal{D}}\left[\log \sigma\left(\left(\sum_{t=0}^{N-1} \beta \log \frac{\pi^*\left(\mathbf{a}_t^w \mid \mathbf{s}_t^w\right)}{\pi_{\mathrm{ref}}\left(\mathbf{a}_t^w \mid \mathbf{s}_t^w\right)}\right)-\left(\sum_{t=0}^{M-1} \beta \log \frac{\pi^*\left(\mathbf{a}_t^l \mid \mathbf{s}_t^l\right)}{\pi_{\mathrm{ref}}\left(\mathbf{a}_t^l \mid \mathbf{s}_t^l\right)}\right)\right)\right]
\end{equation}

\subsection{Proof of Dynamic Token Weight in Token-level DPO}
\label{app: change_beta}

In classic RLHF methods, the optimization objective is typically formulated with an entropy bonus, expressed through a Kullback-Leibler (KL) divergence constraint as follows:

\begin{align}
&
\max_{\pi_\theta} \mathbb{E}_{a_t \sim \pi_\theta(\cdot | \mathbf{s}_t)} \sum_{t=0}^{T} [r(\mathbf{s}_t, \mathbf{a}_t) - \beta \mathcal{D}_{KL}[\pi_{\theta}(\mathbf{a}_t | \mathbf{s}_t)||\pi_{ref}(\mathbf{a}_t | \mathbf{s}_t)]]
\\
&
=\max_{\pi_\theta} \mathbb{E}_{a_t \sim \pi_\theta(\cdot | \mathbf{s}_t)} \sum_{t=0}^{T} [r(\mathbf{s}_t, \mathbf{a}_t) - \beta \log \frac{\pi_{\theta}(\mathbf{a}_t | \mathbf{s}_t)}{\pi_{ref}(\mathbf{a}_t | \mathbf{s}_t)}]
\label{eq: rlhf_objective}
\end{align}

This can be further rewritten by separating the terms involving the reference policy and the entropy of the current policy:

$$\max_{\pi_\theta} \mathbb{E}_{a_t \sim \pi_\theta(\cdot | \mathbf{s}_t)} [ \sum_{t=0}^{T} ( r(\mathbf{s}_t, \mathbf{a}_t) + \beta \log \pi_{ref}(\mathbf{a}_t | \mathbf{s}_t) ) + \beta \mathcal{H}(\pi_\theta) | \mathbf{s}_0 \sim \rho(\mathbf{s}_0) ]$$

When the coefficient $\beta$ is treated as a variable that depends on the timestep $t$ \cite{li20242ddposcalingdirectpreference}, the objective transforms to:

\begin{align}
&
\max_{\pi_\theta} \mathbb{E}_{a_t \sim \pi_\theta(\cdot | \mathbf{s}_t)} \sum_{t=0}^{T} [( r(\mathbf{s}_t, \mathbf{a}_t) + \beta_t \log \pi_{ref}(\mathbf{a}_t | \mathbf{s}_t)) - \beta_t \log \pi_{\theta}(\mathbf{a}_t | \mathbf{s}_t)]
\end{align}

\noindent where $\beta_t$ depends solely on $\mathbf{a}_t$ and $\mathbf{s}_t$. Following the formulation by \citet{Levine2018ReinforcementLA}, the above expression can be recast to incorporate the KL divergence explicitly:

\begin{align}
&
\max_{\pi_\theta} \mathbb{E}_{a_t \sim \pi_\theta(\cdot | \mathbf{s}_t)} \sum_{t=0}^{T} [( r(\mathbf{s}_t, \mathbf{a}_t) + \beta_t \log \pi_{ref}(\mathbf{a}_t | \mathbf{s}_t)) - \beta_t \log \pi_{\theta}(\mathbf{a}_t | \mathbf{s}_t)]
\end{align}

\noindent where the value function  $V(\mathbf{s}_t)$ is defined as:

\begin{align}
V(\mathbf{s}_t) = \beta_t \log \int_{\mathcal{A}} [\exp\frac{r(\mathbf{s}_t, \mathbf{a}_t)}{\beta_t} \pi_{ref}(\mathbf{a}_t | \mathbf{s}_t)] \, d\mathbf{a}_t
\end{align}

When the KL divergence term is minimized—implying that the two distributions are identical—the expectation in Eq. \eqref{eq: rlhf_objective} reaches its maximum value. Therefore, the optimal policy satisfies:

\begin{align}
\pi_\theta(\mathbf{a}_t | \mathbf{s}_t) = \frac{1}{\exp(V(\mathbf{s}_t))} \exp\left(\frac{r(\mathbf{s}_t, \mathbf{a}_t) + \beta_t \log \pi_{ref}(\mathbf{a}_t | \mathbf{s}_t)}{\beta_t}\right)
\end{align}

Based on this relationship, we define the optimal Q-function as:

\begin{equation}
Q^*(s_t, a_t) =
\begin{cases} 
r(s_t, a_t) + \beta_t \log \pi_{ref}(a_t | s_t) + V^*(s_{t+1}), & \text{if } s_{t+1} \text{ is not terminal} \\
r(s_t, a_t) + \beta_t \log \pi_{ref}(a_t | s_t), & \text{if } s_{t+1} \text{ is terminal}
\end{cases}
\label{eq: t_return}
\end{equation}

Consequently, the optimal policy can be expressed as:
\begin{align}
\pi_\theta(\mathbf{a}_t | \mathbf{s}_t) = e^{(Q(\mathbf{s}_t, \mathbf{a}_t) - V(\mathbf{s}_t))/\beta_t}
\label{eq: fixed_point_2}
\end{align}

By taking the natural logarithm of both sides, we obtain a log-linear relationship for the optimal policy at the token level, which is expressed with the optimial Q-function:
\begin{align}
\beta_t \log \pi_\theta(\mathbf{a}_t \mid \mathbf{s}_t) = Q_\theta(\mathbf{s}_t, \mathbf{a}_t) - V_\theta(\mathbf{s}_t)
\end{align}

This equation establishes a direct relationship between the scaled log-ratio of the optimal policy to the reference policy and the reward function $r(\mathbf{s}_t, \mathbf{a}_t)$:

\begin{align}
\beta_t \log \frac{\pi^*(\mathbf{a}_t \mid \mathbf{s}_t)}{\pi_{\text{ref}}(\mathbf{a}_t \mid \mathbf{s}_t)} = r(\mathbf{s}_t, \mathbf{a}_t) + V^*(\mathbf{s}_{t+1}) - V^*(\mathbf{s}_t)
\end{align}

Furthermore, following the definition by \citet{rafailov2024rqlanguagemodel}'s definition, two reward functions $r(\mathbf{s}_t, \mathbf{a}_t)$ and $r'(\mathbf{s}_t, \mathbf{a}_t)$ are considered equivalent if there exists a potential function $\Phi(\mathbf{s})$, such that:

\begin{align}
r'(\mathbf{s}_t, \mathbf{a}_t) =r(\mathbf{s}_t, \mathbf{a}_t) + \Phi(\mathbf{s}_{t+1})  - \Phi(\mathbf{s}_{t})
\end{align}

This equivalence implies that the optimal advantage function remains invariant under such transformations of the reward function. Consequently, we derive why the coefficient $beta$ in direct preference optimization can be variable, depending on the state and action, thereby allowing for more flexible and adaptive policy optimization in RLHF frameworks.

\section{Detailed Experiment Results}
\label{sec:app-results}
In this section, we presented detailed experiment results which are omitted in the main body of this paper due to space limitation. The detailed experiment results of different methods on ComplexBench, FollowBench and AlpacaEval2 are presented in Table \ref{tab:complexbench}, \ref{tab:alpaca-eval} and \ref{tab:followbench}. The detailed results for the ablative studies of confidence metrics is presented in Table \ref{tab:detail-confidence}. The detailed results for the ablative studies of confidence metrics is presented in Table \ref{tab:detail-noising}. We also present a case study in Table \ref{tab:case-study}, which visualize the token-level weight derived from calibrated confidence score.

\begin{table*}[ht]
\centering
\resizebox{1.0\textwidth}{!}{
\begin{tabular}{cc|cccc|cccc}
\hline
\multirow{3}{*}{\textbf{Scenario}} & \multirow{3}{*}{\textbf{Method}} & \multicolumn{8}{c}{\textbf{ComplexBench}}                                                                                                         \\
                                   &                                  & \multicolumn{4}{c}{\textbf{Meta-Llama3-8B-Instruct}}                    & \multicolumn{4}{c}{\textbf{Qwen2-7B-Instruct}}                          \\
                                   &                                  & \textbf{Overall} & \textbf{And}   & \textbf{Chain} & \textbf{Selection} & \textbf{Overall} & \textbf{And}   & \textbf{Chain} & \textbf{Selection} \\ \hline
\multicolumn{2}{c|}{baseline}                          & 61.49            & 57.22          & 57.22          & 53.55              & 67.24            & 62.58          & 62.58          & 58.97              \\ \hline
\multirow{6}{*}{SelfInst}          & Self-Reward                      & 62.45            & 58.23          & 58.23          & 54.07              & 66.98            & 63.02          & 63.02          & 57.75              \\
                                   & w/ BSM                           & 64.13            & 58.01          & 58.01          & 56.62              & 67.02            & 62.37          & 62.37          & 57.85              \\
                                   & w/ GPT-4                         & 64.05            & 59.44          & 59.44          & 54.78              & —                & —              & —              & —                  \\ \cline{2-10} 
                                   & Self-Correct                     & 55.91            & 49.85          & 49.85          & 46.91              & 64.41            & 59.59          & 59.59          & 55.04              \\
                                   & ISHEEP                           & 62.67            & 57.79          & 57.79          & 54.63              & 67.32            & 61.95          & 61.95          & 59.64              \\ \cline{2-10} 
                                   & \textbf{MuSC}                    & \textbf{65.98}   & \textbf{63.45} & \textbf{63.45} & \textbf{55.96}     & \textbf{69.39}   & \textbf{65.45} & \textbf{65.45} & \textbf{59.79}     \\ \hline
\multirow{7}{*}{PreInst}           & Self-Reward                      & 62.03            & 56.94          & 56.94          & 53.09              & 66.45            & 61.37          & 61.37          & 57.64              \\
                                   & w/ BSM                           & 64.30            & 57.58          & 57.58          & 56.47              & 67.43            & 62.95          & 62.95          & 58.41              \\
                                   & w/ GPT-4                         & 63.52            & 59.08          & 59.08          & 53.91              & —                & —              & —              & —                  \\ \cline{2-10} 
                                   & Self-Correct                     & 60.79            & 55.65          & 55.65          & 52.02              & 64.32            & 60.16          & 60.16          & 54.63              \\
                                   & ISHEEP                           & 62.92            & 56.37          & 56.37          & 54.83              & 67.13            & 64.45          & 64.45          & 57.54              \\
                                   & SFT                              & 53.93            & 45.77          & 45.77          & 44.09              & 65.89            & 60.16          & 60.16          & 57.39              \\ \cline{2-10} 
                                   & \textbf{MuSC}                    & \textbf{64.73}   & \textbf{59.23} & \textbf{59.23} & \textbf{55.91}     & \textbf{70.00}   & \textbf{66.88} & \textbf{66.88} & \textbf{61.38}     \\ \hline
\end{tabular}}
\label{tab:complexbench}
\caption{Detailed experiment results of different methods on ComplexBench.}
\label{tab:complexbench}
\end{table*}

\begin{table*}[ht]
\centering
\resizebox{0.75\textwidth}{!}{
\begin{tabular}{cc|ccc|ccc}
\hline
\multirow{3}{*}{\textbf{Scenario}} & \multirow{3}{*}{\textbf{Method}} & \multicolumn{6}{c}{\textbf{FollowBench}}                                                               \\
                                   &                                  & \multicolumn{3}{c}{\textbf{Meta-Llama3-8B-Instruct}} & \multicolumn{3}{c}{\textbf{Qwen2-7B-Instruct}}  \\
                                   &                                  & \textbf{HSR}     & \textbf{SSR}     & \textbf{CSL}   & \textbf{HSR}   & \textbf{SSR}   & \textbf{CSL}  \\ \hline
\multicolumn{2}{c|}{baseline}                                         & 62.39            & 73.07            & 2.76           & 59.81          & 71.69          & 2.46          \\ \hline
\multirow{6}{*}{SelfInst}          & Self-Reward                      & 61.20            & 72.22            & 2.56           & 55.36          & 69.71          & 2.34          \\
                                   & w/ BSM                           & 64.30            & 73.84            & 2.80           & 57.83          & 70.53          & 2.41          \\
                                   & w/ GPT-4                         & 62.18            & 73.34            & 2.66           & —              & —              & —             \\ \cline{2-8} 
                                   & Self-Correct                     & 54.38            & 67.19            & 2.02           & 51.98          & 67.89          & 2.16          \\
                                   & ISHEEP                           & 62.77            & 72.86            & 2.52           & 57.01          & 69.88          & 2.36          \\ \cline{2-8} 
                                   & \textbf{MuSC}                    & \textbf{66.71}   & \textbf{74.84}   & \textbf{2.92}  & \textbf{62.60} & \textbf{72.57} & \textbf{2.82} \\ \hline
\multirow{7}{*}{PreInst}           & Self-Reward                      & 60.88            & 72.17            & 2.64           & 56.45          & 70.00          & 2.44          \\
                                   & w/ BSM                           & 63.96            & 73.78            & 2.66           & 58.02          & 70.62          & 2.42          \\
                                   & w/ GPT-4                         & 64.02            & 73.26            & 2.64           & —              & —              & —             \\ \cline{2-8} 
                                   & Self-Correct                     & 60.11            & 70.94            & 2.70           & 49.47          & 66.35          & 1.98          \\
                                   & ISHEEP                           & 63.54            & 73.21            & 2.64           & 55.52          & 69.62          & 2.28          \\
                                   & SFT                              & 50.06            & 66.48            & 2.04           & 47.36          & 64.67          & 1.96          \\ \cline{2-8} 
                                   & \textbf{MuSC}                    & \textbf{66.90}   & \textbf{75.11}   & \textbf{2.99}  & \textbf{62.73} & \textbf{73.09} & \textbf{2.86} \\ \hline
\end{tabular}}
\caption{Detailed experiment results of different methods on FollowBench.}
\label{tab:followbench}
\end{table*}

\begin{table*}[ht]
\centering
\resizebox{0.9\textwidth}{!}{
\begin{tabular}{cc|cccccc}
\hline
\multirow{3}{*}{\textbf{Scenario}} & \multirow{3}{*}{\textbf{Method}} & \multicolumn{6}{c}{\textbf{AlpacaEval2}}                                                                          \\
                                   &                                  & \multicolumn{3}{c}{\textbf{Meta-Llama3-8B-Instruct}}    & \multicolumn{3}{c}{\textbf{Qwen2-7B-Instruct}}          \\
                                   &                                  & \textbf{LC (\%)} & \textbf{WR (\%)} & \textbf{Avg. Len} & \textbf{LC (\%)} & \textbf{WR (\%)} & \textbf{Avg. Len} \\ \hline
\multicolumn{2}{c|}{baseline}                                         & 21.07            & 18.73            & 1702              & 15.53            & 13.70            & 1688              \\ \hline
\multirow{6}{*}{SelfInst}          & Self-Reward                      & 19.21            & 19.18            & 1824              & 16.81            & 15.66            & 1756              \\
                                   & w/ BSM                           & 19.03            & 18.34            & 1787              & 16.94            & 15.09            & 1710              \\
                                   & w/ GPT-4                         & 19.55            & 18.53            & 1767              & —                & —                & —                 \\ \cline{2-8} 
                                   & Self-Correct                     & 7.97             & 9.34             & 1919              & 14.01            & 10.92            & 1497              \\
                                   & ISHEEP                           & 22.00            & 19.50            & 1707              & 16.99            & 14.04            & 1619              \\ \cline{2-8} 
                                   & \textbf{MuSC}                    & \textbf{23.87}   & \textbf{20.91}   & \textbf{1708}     & \textbf{20.08}   & \textbf{15.67}   & \textbf{1595}     \\ \hline
\multirow{7}{*}{PreInst}           & Self-Reward                      & 19.93            & 19.04            & 1789              & 15.98            & 15.62            & 1796              \\
                                   & w/ BSM                           & 20.98            & 20.75            & 1829              & 17.17            & 16.21            & 1764              \\
                                   & w/ GPT-4                         & 18.02            & 17.74            & 1804              & —                & —                & —                 \\ \cline{2-8} 
                                   & Self-Correct                     & 6.20             & 5.81             & 1593              & 14.46            & 14.02            & 1737              \\
                                   & ISHEEP                           & 20.23            & 17.86            & 1703              & 16.52            & 13.36            & 1627              \\
                                   & SFT                              & 10.00            & 6.22             & 1079              & 9.52             & 5.25             & 979               \\ \cline{2-8} 
                                   & \textbf{MuSC}                    & \textbf{23.74}   & \textbf{19.53}   & \textbf{1631}     & \textbf{20.29}   & \textbf{15.91}   & \textbf{1613}     \\ \hline
\end{tabular}}
\caption{Detailed experiment results of different methods on AlpacaEval2.}
\label{tab:alpaca-eval}
\end{table*}

\begin{table}[ht]
\centering
\resizebox{0.95\textwidth}{!}{
\begin{tabular}{cc|ccccc|ccccc}
\toprule
\multirow{3}{*}{\textbf{Scenario}} & \multirow{3}{*}{\textbf{Method}} & \multicolumn{5}{c|}{\textbf{Meta-Llama-3-8B-Instruct}}                                    & \multicolumn{5}{c}{\textbf{Qwen-2-7B-Instruct}}                                          \\
                                   &                                  & \multicolumn{3}{c}{\textbf{CF-Bench}}         & \multicolumn{2}{c|}{\textbf{AlpacaEval2}} & \multicolumn{3}{c}{\textbf{CF-Bench}}         & \multicolumn{2}{c}{\textbf{AlpacaEval2}} \\
                                   &                                  & \textbf{CSR}  & \textbf{ISR}  & \textbf{PSR}  & \textbf{LC (\%)}   & \textbf{Avg. Len}       & \textbf{CSR}  & \textbf{ISR}  & \textbf{PSR}  & \textbf{LC (\%)}   & \textbf{Avg. Len}      \\ \midrule
\multirow{6}{*}{PreInst}           & Baseline                         & 0.64          & 0.24          & 0.34          & 21.07                & 1702               & 0.74          & 0.36          & 0.49          & 15.53                & 1688              \\ \cline{2-12} 
                                   & w/ perplexity                    & 0.70          & 0.32          & 0.43          & 22.99                & 1744               & 0.79          & 0.43          & 0.54          & 19.31                & 1675              \\
                                   & w/ PMI                           & 0.69          & 0.29          & 0.41          & 21.92                & 1713               & 0.78          & 0.43          & 0.55          & 17.42                & 1651              \\
                                   & w/ KLDiv                         & 0.69          & 0.31          & 0.42          & 21.86                & 1686               & 0.78          & 0.42          & 0.54          & 18.73                & 1614              \\
                                   & \textbf{w/ entropy}                       & \textbf{0.71} & \textbf{0.34} & \textbf{0.44} & \textbf{23.74}       & 1631               & \textbf{0.80} & \textbf{0.46} & \textbf{0.58} & \textbf{20.38}       & 1683              \\ \bottomrule
\end{tabular}}
\caption{Detailed experiment results of different confidence metrics as the fine-grained weight.}
\label{tab:detail-confidence}
\end{table}

\setlength{\fboxsep}{0pt}

\begin{table}[t]
\centering
\resizebox{1.0\textwidth}{!}{
\begin{tabular}{l}
\hline
\textbf{Instruction 1}: Identify the incorrect word in the given sentence: "The group acquired the skill to ski\\ in a suitable manner." Rewrite the identified incorrect word correctly, ensuring the new sentence\\ adheres to proper capitalization, punctuation, and grammar rules. Include at least one synonym of\\ the incorrect word to test the model's understanding of context and vocabulary, and ensure the synon-\\ ym is used appropriately in the context of the sentence. \\\hline
\textbf{Chosen Response 1}: \colorbox{red!38}{\strut \:The}\colorbox{red!30}{\strut \:group}\colorbox{red!12}{\strut \:acquired}\colorbox{red!50}{\strut \:the}\colorbox{red!25}{\strut \:skill}\colorbox{red!50}{\strut \:to}\colorbox{red!38}{\strut \:ski}\colorbox{red!50}{\strut \:in}\colorbox{red!50}{\strut \:a}\colorbox{red!0}{\strut \:proficient}\colorbox{red!25}{\strut \:manner}\colorbox{red!50}{\strut \:.}\colorbox{red!35}{\strut \:The}\colorbox{red!12}{\strut \:incorrect}\colorbox{red!25}{\strut \:word}\\\colorbox{red!50}{\strut \:is}\colorbox{red!38}{\strut \:"suitable"}\colorbox{red!50}{\strut \:.}\colorbox{red!45}{\strut \:A}\colorbox{red!18}{\strut \:synonym}\colorbox{red!50}{\strut \:for}\colorbox{red!38}{\strut \:"suitable"}\colorbox{red!50}{\strut \:is}\colorbox{red!5}{\strut \:"appropriate"}\colorbox{red!50}{\strut \:,}\colorbox{red!40}{\strut \:but}\colorbox{red!50}{\strut \:it}\colorbox{red!25}{\strut \:doesn't}\colorbox{red!10}{\strut \:fit}\colorbox{red!50}{\strut \:the}\colorbox{red!32}{\strut \:context}\colorbox{red!50}{\strut \:of}\colorbox{red!50}{\strut \:the}\colorbox{red!28}{\strut \:sentence}\colorbox{red!50}{\strut \:.}\\\colorbox{red!8}{\strut \:"Proficient"}\colorbox{red!50}{\strut \:is}\colorbox{red!50}{\strut \:a}\colorbox{red!18}{\strut \:better}\colorbox{red!20}{\strut \:synonym}\colorbox{red!50}{\strut \:in}\colorbox{red!45}{\strut \:this}\colorbox{red!25}{\strut \:context}\colorbox{red!50}{\strut \:as}\colorbox{red!45}{\strut \:it}\colorbox{red!50}{\strut \:means}\colorbox{red!5}{\strut \:skilled}\colorbox{red!50}{\strut \:or}\colorbox{red!32}{\strut \:expert}\colorbox{red!50}{\strut \:.} \\\hline

\textbf{Rejected Response 1}: \colorbox{red!14}{\strut \:The}\colorbox{red!25}{\strut \:word}\colorbox{red!43}{\strut \:"acquired"}\colorbox{red!32}{\strut \:seems}\colorbox{red!36}{\strut \:incorrect}\colorbox{red!0}{\strut \:in}\colorbox{red!0}{\strut \:the}\colorbox{red!27}{\strut \:given}\colorbox{red!23}{\strut \:sentence}\colorbox{red!0}{\strut \:.}\colorbox{red!0}{\strut \:A}\colorbox{red!39}{\strut \:better}\colorbox{red!20}{\strut \:word}\colorbox{red!0}{\strut \:to}\\\colorbox{red!16}{\strut \:use}\colorbox{red!0}{\strut \:in}\colorbox{red!0}{\strut \:this}\colorbox{red!27}{\strut \:context}\colorbox{red!23}{\strut \:would}\colorbox{red!0}{\strut \:be} \colorbox{red!50}{\strut \:"learned"}\colorbox{red!0}{\strut \:.}\colorbox{red!18}{\strut \:Rewritten}\colorbox{red!20}{\strut \:sentence}\colorbox{red!0}{\strut \::}\colorbox{red!9}{\strut \:"The}\colorbox{red!23}{\strut \:group}\colorbox{red!30}{\strut \:learned}\colorbox{red!0}{\strut \:the}\colorbox{red!25}{\strut \:skill}\colorbox{red!0}{\strut \:to}\colorbox{red!9}{\strut \:ski}\colorbox{red!0}{\strut \:in}\colorbox{red!0}{\strut \:a}\\\colorbox{red!36}{\strut \:suitable}\colorbox{red!23}{\strut \:manner}\colorbox{red!0}{\strut \:."} \\\hline




\textbf{Instruction 2}: Rewrite the given sentence in C\# syntax, using an idiom or metaphor to express dis-\\ belief, and ensure the sentence maintains its original meaning while being grammatically correct in\\ C\#. The sentence to be rewritten is: "The idea that pigs can fly seems more plausible than his expla-\\nation for being late."\\\hline

\textbf{Chosen Response 2}:\colorbox{red!50}{\strut \:csharp\textbackslash n}\colorbox{red!50}{\strut \:using}\colorbox{red!50}{\strut \:System}\colorbox{red!50}{\strut \:;\textbackslash n}\colorbox{red!50}{\strut \:class}\colorbox{red!44}{\strut \:Program\textbackslash n}\colorbox{red!50}{\strut \:\{\textbackslash n}\colorbox{red!50}{\strut \:static}\colorbox{red!50}{\strut \:void}\colorbox{red!44}{\strut \:Main}\colorbox{red!50}{\strut \:(}\colorbox{red!50}{\strut \:string}\colorbox{red!39}{\strut \:[]}\colorbox{red!50}{\strut \:args}\\\colorbox{red!50}{\strut \:\textbackslash n)}\colorbox{red!44}{\strut \:\{\textbackslash n}\colorbox{red!50}{\strut \:Console.WriteLine}\colorbox{red!50}{\strut \:(}\colorbox{red!36}{\strut \:"His}\colorbox{red!19}{\strut \:explanation}\colorbox{red!50}{\strut \:for}\colorbox{red!50}{\strut \:being}\colorbox{red!36}{\strut \:late}\colorbox{red!50}{\strut \:is}\colorbox{red!50}{\strut \:so}\colorbox{red!22}{\strut \:baffling}\colorbox{red!50}{\strut \:that}\colorbox{red!31}{\strut \:it's}\colorbox{red!0}{\strut \:akin}\colorbox{red!50}{\strut \:to}\colorbox{red!31}{\strut \:saying}\colorbox{red!39}{\strut \:pigs}\\\colorbox{red!50}{\strut \:can}\colorbox{red!36}{\strut \:fly}\colorbox{red!50}{\strut \:,}\colorbox{red!19}{\strut \:making}\colorbox{red!50}{\strut \:it}\colorbox{red!33}{\strut \:more}\colorbox{red!8}{\strut \:believable}\colorbox{red!50}{\strut \:in}\colorbox{red!25}{\strut \:comparison}\colorbox{red!44}{\strut \:.}\colorbox{red!50}{\strut \:)}\colorbox{red!50}{\strut \:;\textbackslash n}\colorbox{red!50}{\strut \:\}\textbackslash n}\colorbox{red!50}{\strut \:\}} \\\hline
\textbf{Rejected Response 2}:\colorbox{red!50}{\strut \:Wow}\colorbox{red!0}{\strut \:,}\colorbox{red!15}{\strut \:it's}\colorbox{red!30}{\strut \:truly}\colorbox{red!39}{\strut \:hard}\colorbox{red!0}{\strut \:to}\colorbox{red!20}{\strut \:believe}\colorbox{red!0}{\strut \:that}\colorbox{red!9}{\strut \:pigs}\colorbox{red!33}{\strut \:flying}\colorbox{red!20}{\strut \:could}\colorbox{red!0}{\strut \:be}\colorbox{red!11}{\strut \:more}\colorbox{red!39}{\strut \:plausible}\colorbox{red!0}{\strut \:than}\\\colorbox{red!15}{\strut \:my}\colorbox{red!22}{\strut \:explanation}\colorbox{red!4}{\strut \:for}\colorbox{red!0}{\strut \:being}\colorbox{red!11}{\strut \:late}\colorbox{red!4}{\strut \:!}\\\hline

\end{tabular}}
\caption{Visualization of dynamic weights derived for chosen and rejected responses, based on our proposed calibrated entropy score. We select two samples from the datasets as an illustration.}
\label{tab:case-study}
\end{table}

